\documentclass[lettersize,journal]{IEEEtran}

\usepackage{amsmath,amsfonts}
\usepackage{algorithmic}
\usepackage{algorithm}
\usepackage{array}
\usepackage{textcomp}
\usepackage{stfloats}
\usepackage{url}
\usepackage{verbatim}
\usepackage{graphicx}
\usepackage{cite}
\usepackage[dvipsnames,table,xcdraw]{xcolor}
\usepackage{amsmath}
\usepackage{amsfonts}
\usepackage{nicefrac}
\usepackage{comment}
\usepackage{hyperref}
\usepackage{xspace}
\usepackage{multirow}
\usepackage{subcaption}
\usepackage{graphbox}
\usepackage[normalem]{ulem}
\useunder{\uline}{\ul}{}

\usepackage{makecell}
\usepackage{graphicx}
\usepackage{nicematrix}

\usepackage{enumitem}  

\usepackage{amssymb}
\usepackage{pifont}
\usepackage{dsfont} 
\usepackage{balance}
\newcommand{\cmark}{\ding{51}}%
\newcommand{\xmark}{\ding{55}}%

\DeclareMathOperator*{\argmin}{arg\,min}
\DeclareMathOperator*{\argmax}{arg\,max}

\newcommand{\bm}{\boldsymbol{m}}
\newcommand{\bmhat}{\boldsymbol{\hat{m}}}
\newcommand{\bM}{\boldsymbol{M}}
\newcommand{\bMhat}{\boldsymbol{\widehat{M}}}
\newcommand{\bO}{\mathbf{O}}

\newcommand{\bp}{\boldsymbol{p}}
\newcommand{\bP}{\mathbf{P}}

\newcommand{\bPhat}{\mathbf{\widehat{P}}}

\newcommand{\bs}{\boldsymbol{s}}
\newcommand{\bS}{\mathbf{S}}
\newcommand{\bshat}{\boldsymbol{\hat{s}}}
\newcommand{\bShat}{\mathbf{\widehat{S}}}

\newcommand{\by}{\boldsymbol{y}}
\newcommand{\bY}{\mathbf{Y}}

\newcommand{\mL}{\mathcal{L}}

\makeatletter
\DeclareRobustCommand\onedot{\futurelet\@let@token\@onedot}
\def\@onedot{\ifx\@let@token.\else.\null\fi\xspace}
\def\eg{\emph{e.g}\onedot} 
\def\ie{\emph{i.e}\onedot}

\def\etal{\emph{et al}\onedot}
\makeatother

\usepackage{xspace} 
\newcommand{\ours}{MaskPlanner\xspace}

\usepackage{bbm}

\begin{document}

\title{MaskPlanner: Learning-Based Object-Centric Motion Generation from 3D Point Clouds}

\author{Gabriele Tiboni$^{1}$, Raffaello Camoriano$^{1, 2}$, Tatiana Tommasi$^{1}$
\thanks{$^{1}$Department of Control and Computer Engineering, Politecnico di Torino, Turin, Italy {\tt\small first.last@polito.it}}%
\thanks{$^{2}$Istituto Italiano di Tecnologia, Genoa, Italy}%
}

\markboth{Under review}{}

\maketitle

\begin{abstract}
Object-Centric Motion Generation (OCMG) plays a key role in a variety of industrial applications---such as robotic spray painting and welding---requiring efficient, scalable, and generalizable algorithms to plan multiple long-horizon trajectories over free-form 3D objects.
However, existing solutions rely on specialized heuristics, expensive optimization routines, or restrictive geometry assumptions that limit their adaptability to real-world scenarios.
In this work, we introduce a novel, fully data-driven framework that tackles OCMG directly from 3D point clouds, learning to generalize expert path patterns across free-form surfaces.
We propose \emph{MaskPlanner}, a deep learning method that predicts local path segments for a given object while simultaneously inferring ``path masks" to group these segments into distinct paths.
This design induces the network to capture both local geometric patterns and global task requirements in a single forward pass.
Extensive experimentation on a realistic robotic spray painting scenario shows that our approach attains near-complete coverage (above 99\%) for unseen objects, while it remains task-agnostic and does not explicitly optimize for paint deposition.
Moreover, our real-world validation on a 6-DoF specialized painting robot demonstrates that the generated trajectories are directly executable and yield expert-level painting quality.
Our findings crucially highlight the potential of the proposed learning method for OCMG to reduce engineering overhead and seamlessly adapt to several industrial use cases.
\end{abstract}

\section{Introduction}
\label{sec:introduction}
\IEEEPARstart{M}{otion} generation conditioned on 3D objects is fundamental to numerous industrial robotic applications, including spray painting, welding, sanding, and cleaning.
Despite the different objectives, these tasks share key challenges arising from the complexity of free-form 3D inputs and the high-dimensional outputs required to define complete robot programs.
In particular, the offline generation of long-horizon motions demands substantial computational resources for both optimization and planning.
Additionally, encoding human expert behavior into explicit optimization objectives remains extremely challenging for such complex tasks.
To address these difficulties, robotics practitioners often rely on task-specific knowledge, impose strong simplifying assumptions regarding the object shapes, and develop heuristic algorithms to render each task individually more tractable.
However, these solutions necessitate extensive re-engineering for each new product, making the process time-consuming, costly, and unable to efficiently adapt to new scenarios. 
In this context, establishing a unifying paradigm to tackle these tasks is a crucial step for the community to transition from handcrafted strategies to techniques that are scalable and capable of generalization.

\begin{figure}[t]
    \centering
    \includegraphics[width=\linewidth]{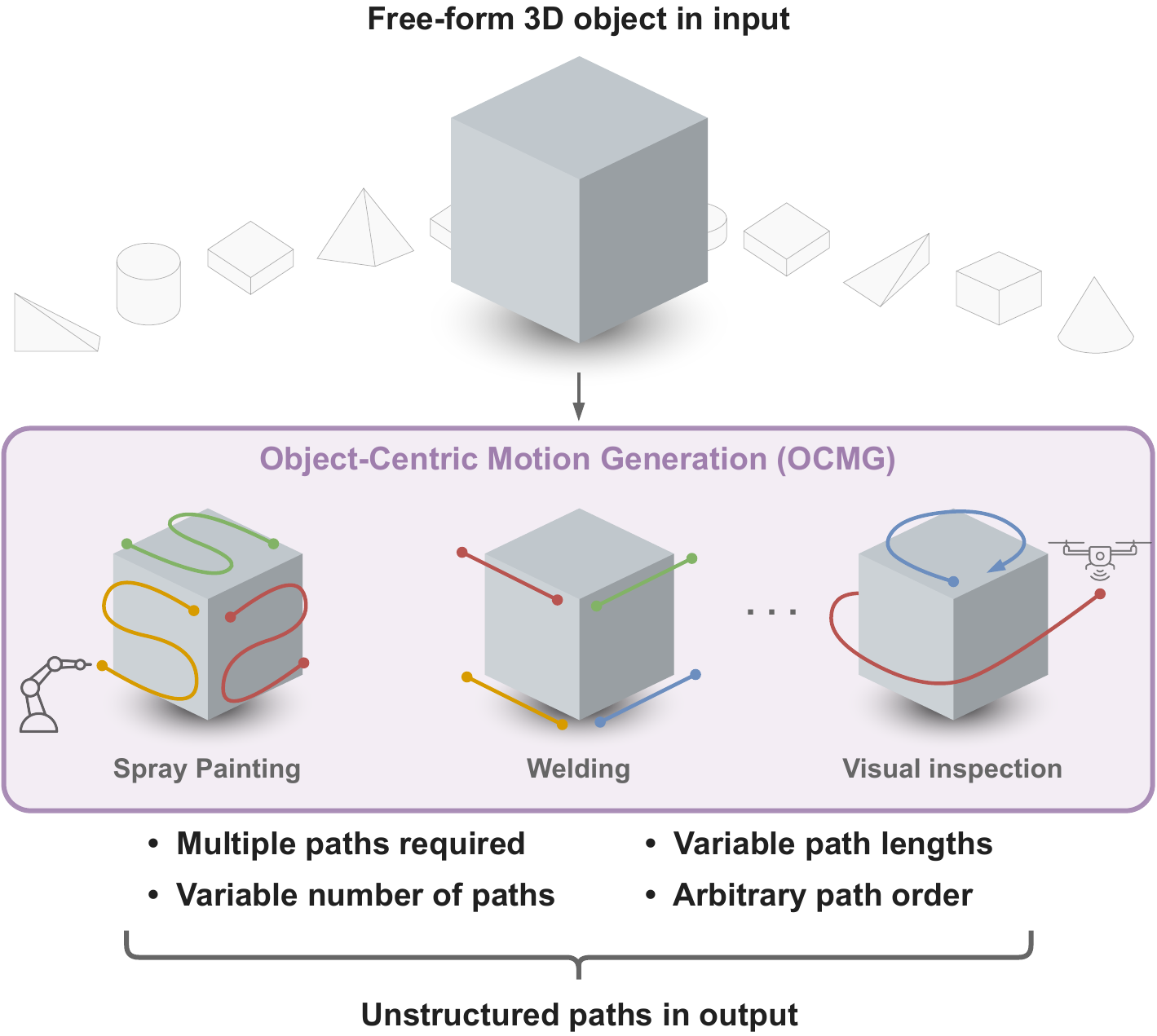}
    \caption{Several object-centric robotic applications may be unified under a single problem formulation, as they share common assumptions on the desired output paths---referred to as unstructured paths.}
    \vspace{-6pt}
    \label{fig:intro_figure}
\end{figure}

To this end, we formalize the \emph{Object-Centric Motion Generation} (OCMG) framework, a novel problem setting that unifies robotic tasks aiming to generate \emph{multiple}, \emph{long-horizon} paths based on \emph{static}, \emph{free-form} 3D objects.
For the sake of a general formulation, we consider output paths to be \emph{unstructured}: no predefined path order is assumed, and paths can generally vary in number and length according to the input object.
Notably, these properties are shared by a wide range of robotic applications---as illustrated in Figure~\ref{fig:intro_figure}---where negligible dynamic interactions with the 3D objects are involved
and global object geometric information is given.

Robotic spray painting is an industrially-relevant example of the OCMG problem:
multiple paths are necessary to paint a single object, and the resulting paint coverage is invariant with respect to the order of execution---\eg, separate paths can be executed in parallel across multiple robots.
Furthermore, motion generation can occur offline, and no reactive planning is needed since global information of the surface geometry is available.
Note that the pattern of the spray painting paths varies significantly with each object instance, making it difficult to codify general rules---\eg, human experts in the field rely on sophisticated, high-level reasoning and costly trial-and-error to determine robot programs based on the 3D geometry of the objects.
Existing research studies resort to decoupling the spray painting task in (i)~3D object partitioning into convex surfaces, and (ii)~offline trajectory optimization through either domain-specific heuristics~\cite{Sheng_Automated_2000,Chen_Automated_2008,Li_Automatic_2010,Andulkar_Incremental_2015,atkar24uniform,gleeson2022generating}, or reinforcement learning-based policies~\cite{Kiemel_PaintRL_2019}.
Yet, such approaches make simplified premises on the structure of output trajectories, require expensive optimization routines, and are heavily tailored to specific shapes and convex surfaces only.
These issues leave robotic spray painting solutions largely limited in flexibility and generalization capabilities, despite their relevance in product manufacturing.

Recently, interest in purely data-driven approaches has grown for extrapolating path patterns without the need to explicitly encode optimization objectives and task-specific constraints.
Assuming that expert data is available, learning-based methods pave the way for solutions that are scalable, cheap to deploy at inference time, and generalizable to unseen scenarios.
A number of successful applications demonstrate the potential of data-driven methods in related problems, such as motion forecasting in autonomous driving~\cite{yuan2021agentformer,ngiam2022scenetr,varadarajan2022multipath++,nayakanti2023wayformer}, multi-agent imitation learning~\cite{SRINIVASAN2021598}, or socially-compliant robot navigation~\cite{kretzschmar2013featuretrajpred,kretzschmar2014learningsocialnavigation,pfeiffer2016predicting}.
Yet, these solutions consider a short prediction horizon, assume a fixed number of paths, and do not deal with 3D input data.
Tiboni \etal~\cite{tiboni2023paintnet} recently proposed the first approach to handle unstructured paths conditioned on 3D objects, predicting disconnected path segments across the object surface rather than directly inferring long-horizon paths. 
This approach allows for accurate local predictions, but crucially lacks a way to organize predicted segments into separate paths, and finally concatenate them to generate long-horizon paths.

In this work, we propose \emph{\ours}, a novel deep learning method to address OCMG tasks directly from expert data, building on~\cite{tiboni2023paintnet}.
Our pipeline breaks down the motion generation problem into the joint prediction of (1)~path segments and (2)~path masks, in a single forward pass. Particularly, we propose learning binary masks over predicted segments to identify which path each segment belongs to.
This strategy allows the network to simultaneously make local (segment predictions) and global (mask predictions) planning decisions in one step.
In turn, our method effectively infers the required \emph{number} of paths and the \emph{length} of each path according to a given input object.

When tested in the field of robotic spray painting, \ours is capable of predicting segments and masks for 40 paths in only 100ms, spanning a total length of 70 meters and 8 minutes of execution time, and achieving near-optimal paint coverage on held-out 3D objects in simulation.
Moreover, we successfully execute the generated paths on a real 6-DoF specialized painting robot for previously unseen object instances, achieving qualitative results that are indistinguishable from those obtained via human-expert trajectories. 
Overall, we make significant progress in addressing the OCMG problem, and focus on spray painting as a representative application to conduct a thorough experimental evaluation.\\

Our novel contributions can be summarized as follows:
\begin{itemize}[leftmargin=*,itemsep=2pt]
    \item \textbf{Mask predictions}:
        we propose \ours, a novel deep learning method that predicts path segments along with a set of masks, identifying which segments belong to the same path.
    \item \textbf{Improved segment predictions}:
        we formalize a multi-component loss function based on the Chamfer Distance and tailored to segment prediction. We study the effect of each component with an extensive ablation analysis.
    \item \textbf{Segment concatenation}:
        a novel post-processing step is designed to filter and concatenate the segments clustered within the same mask into long-horizon paths.
    \item \textbf{Improved benchmarking}:
        we release a new public dataset extending that in~\cite{tiboni2023paintnet} by more than three times in size.
        Two novel baselines are implemented ad-hoc for comparison with methods that perform na\"ive autoregressive or one-shot predictions for object-centric motion generation. 
    \item \textbf{Real-world validation}:
        we assess the performance of \ours by executing the predicted paths on a 6-DoF spray painting robot, achieving expert-level painting quality on previously unseen object instances. 
\end{itemize}

\section{Related Work}
\label{sec:related_work}
In this section, we provide a review of the literature related to the OCMG problem (Sec.~\ref{sec:motion_plan_gen}-\ref{sec:3dpointclouds}) with an in-depth focus on previous works in robotic spray painting (Sec.~\ref{sec:painting}). 

\subsection{Learning-based Motion Planning and Generation}
\label{sec:motion_plan_gen}
\emph{Motion planning} involves finding low-cost, goal-conditioned trajectories in a given environment while accounting for task-specific constraints---such as enforcing kinematic or dynamic feasibility, safety, or smoothness~\cite{karur2021surveypathplanning}.
Conventional methods based on search or sampling over a discrete representation of the state space tend to be computationally expensive~\cite{kicki2023fasttateo,ichnowski2020deepmotionplanning}, hence unsuitable for real-world applications where planning has tight time requirements or spans high-dimensional spaces. 
Learning-based approaches for motion planning have been proposed to speed up planning times~\cite{surveylearningrobotmotionplanning2021} by predicting the sampling distribution for sampling-based methods~\cite{wang2020neuralrrt,cheng2020learningpathplanning}, warm-starting traditional solvers~\cite{ichnowski2020deepmotionplanning}, promoting the feasibility of planned trajectories~\cite{kicki2023fasttateo}, performing end-to-end planning~\cite{pfeiffer2017perception,bency2019neuraloraclenet}, or training Reinforcement Learning policies~\cite{tsounis2020deepgaitrl,kim2020motion} to predict short-term actions that maximize cumulative rewards.
Motion planning solutions yet focus on reaching a specified goal state from a predefined start location, hence they are not designed to generate complex path patterns that mimic expert behavior. 

\emph{Motion generation}~\cite{bekris2024motiongensurvey} encompasses a broader scope than traditional planning, as it may not involve start and goal states and often necessitates adherence to task-specific motion patterns. 
Among learning-based approaches, Sasagawa et al.~\cite{sasagawa2021motionbilateral} train a recurrent neural network to tackle long-term motion generation in complex tasks such as writing letters, which requires separate sequential strokes.
Saito et al.~\cite{saito2023structured} adopt supervised deep learning to tackle long-horizon manipulation tasks 
and breaking the motion generation problem down into subgoals prediction. 
Neural networks also proved effective in generating human-like whole-body trajectories by learning from human motion capture data, in both humanoid robotics~\cite{viceconte2022adherent} and character control~\cite{zhang2018mode}.
%
%
Imitation Learning~(IL)~\cite{ross2011reductionbc,behav_cloning,ho2016gail,panilautonomousdriving,Ze2024DP3diffusionpolicy} tackles motion generation assuming that a reward function is described implicitly through expert demonstrations, hence solving the task by learning from data.
In particular, Behavioral Cloning (BC)~\cite{ross2011reductionbc,behav_cloning} consists of supervised learning techniques that directly find a mapping from the current state to the optimal action, \eg., through regression methods.
Alternatively, Inverse Reinforcement Learning (IRL)~\cite{abbeel2004apprenticeshipirl,ziebart2008maximumirl} aims at learning a representation of the underlying reward function the human experts used to generate their actions. IRL has been successfully deployed to learn parking lot navigation strategies~\cite{abbeel2008apprenticeshipirlplanning}, human-like driving behavior~\cite{wulfmeier2016watchirlpathplanning}, and long-term motion forecasting~\cite{shkurti2018modelpursuit}. 

Notably, BC and IRL methods typically frame motion generation as a sequential decision making problem in an unknown dynamic environment with the Markov property---a Markov Decision Process. While our work is also fully data-driven, we address the challenge of global, long-horizon motion generation with complete state information.
Furthermore, although adaptations of IL to multiple agents~\cite{SRINIVASAN2021598} and global trajectory learning~\cite{osa2017guiding,duan2024structured,behav_cloning} were proposed, no method can manage unstructured paths---namely, they fail to model scenarios where the number of agents/paths is unknown, and no temporal correlation exists among separate paths.
Finally, our work focuses on learning paths that are generalizable across complex 3D shapes directly, a setting which has been rarely addressed before in robot imitation learning~\cite{schulman2016,Ze2024DP3diffusionpolicy}.

\subsection{Motion Prediction}
Motion prediction aims at anticipating the motion of multiple agents ahead in the future.
To solve this problem, existing methods generally employ supervised learning techniques from observed trajectories, with applications to autonomous driving~\cite{yuan2021agentformer,ngiam2022scenetr,varadarajan2022multipath++,nayakanti2023wayformer}, human motion forecasting~\cite{alahi2016social,gupta2018social,rudenko2020humanpredictionsurvey}, and socially-compliant robot navigation~\cite{kretzschmar2013featuretrajpred,kretzschmar2014learningsocialnavigation,pfeiffer2016predicting}. 
Pfeiffer et al.~\cite{pfeiffer2016predicting} leverage the maximum entropy principle to learn a joint probability distribution over the future trajectories of all agents in the scene from data, including the controllable robot.
Gupta et al.~\cite{gupta2018social} predict socially-plausible human motion paths using a recurrent model and generative adversarial networks.
Nayakanti et al.~\cite{nayakanti2023wayformer} adopt a family of attention-based networks for motion forecasting in autonomous driving, investigating the most effective ways to fuse scene information including agents'~history, road configuration, and traffic light state.

Notably, motion prediction deals with output paths that are jointly executed through decentralized agents that move simultaneously over time, from a given starting state.
In turn, numerous strategies were proposed to aggregate information across agents and across time, such as
pooling layers~\cite{alahi2016social,gupta2018social}, independent self-attention for each axis~\cite{yu2020spatiotemporal}, or joint attention mechanisms on multiple axes~\cite{yuan2021agentformer,ngiam2022scenetr,nayakanti2023wayformer}.
In contrast, the OCMG problem considers learning a set of disjoint paths that are uncorrelated in time---\eg., they may be executed separately at different times, in an arbitrary order, and from unknown starting states.
In addition, the motion prediction literature assumes a known, fixed number of agents in the scene. Recently, Gu et al.~\cite{gu2023vip3d} introduced the first end-to-end approach to couple motion prediction with object detection and tracking, effectively dealing with a varying number of agents that is automatically inferred at test time.
Yet, adapting these works to heterogeneous path lengths and long-horizon motions is an open problem---the literature focuses on fixed prediction horizons of only 3-8 seconds.
Overall, the temporal correlation of predicted paths and assumptions on fixed, short-horizon forecasting render motion prediction methods unsuitable for direct application to the OCMG setting.

\subsection{Set Prediction}
Canonical deep learning models are not designed to directly predict sets, \ie collections of permutation-invariant elements with varying cardinality.
Early works addressed this issue in the context of multi-label classification~\cite{rezatofighi2017deepsetnet}, where an unknown number of labels must be associated with a given input. 
For regression tasks of set prediction (\eg, Object Detection), the difficulty is instead to avoid generating near-duplicate outputs (\ie, near-identical bounding boxes) due to an unknown number of output elements.
This was originally mitigated via postprocessing techniques such as non-maximal suppression~\cite{erhan2014scalable,redmon2016yolo}. 
Later, auto-regressive recurrent models were proposed for sequentially predicting output sets~\cite{vinyalsseqtoseq,stewart2016end}, but these approaches were eventually outperformed by transformer-based architectures~\cite{carion2020detr,cheng2021maskformer}.
Transformers excel in such tasks by leveraging attention mechanisms to decode output elements in parallel and capture long-range dependencies, resulting in more robust and accurate set predictions.

Regardless of the architecture, the loss function designed for set prediction must always be invariant by a permutation of the predictions, or the ground truth.
This can be achieved by \emph{matching} predictions with ground truths before the loss computation, either implicitly, leveraging on a moving window across the input image~\cite{redmon2016yolo}, or explicitly, by solving a bipartite matching problem~\cite{erhan2014scalable,stewart2016end,cheng2021maskformer}.
The latter approach has been widely adopted in Object Detection~\cite{carion2020detr} and Panoptic Segmentation~\cite{cheng2021maskformer} tasks using the Hungarian algorithm~\cite{kuhn1955hungarian}.

\subsection{3D Deep Learning from Point Clouds}
\label{sec:3dpointclouds}
3D deep learning architectures apply predictive models to process free-form 3D data~\cite{ahmed2018survey}, typically represented as voxel grids, meshes, or point clouds.
Particularly, point cloud representations describe objects as unstructured sets of 3D points, and were successfully proposed to perform tasks such as 3D object classification~\cite{Qi_Pointnet_2017} and segmentation~\cite{Qi_Pointnet++_2017}, and shape completion~\cite{Yuan_Pcn_2018,Alliegro_Denoise_2021}.
The latter task involves reconstructing missing parts of a 3D object or scene from incomplete input data and has shown to be relevant for robotics applications \cite{completionHumanoids,3dsgrasp}.
Similarly to the OCMG framework, the input is a free-form 3D shape and the output is unstructured, \ie, the unordered set of points that fill missing input regions.
Inspired by these methods, our work leverages the expressive power of 3D deep learning architectures and adapts them to predict unstructured robotic paths that generalize to new object instances.

\begin{table*}[t]
\centering
\caption{Literature review with a selected number of exemplary works in fields of applications related to OCMG. We dissect each work to shed light on the differences and similarities with our problem setting.}
\label{tab:literature_review}
\def\arraystretch{1.25}%
\resizebox{\linewidth}{!}{
\begin{NiceTabular}{|c|c|c|c|c|c|c|c|c|c|l|}
\CodeBefore
  \rectanglecolor[gray]{0.9}{1-12}{2-0}
\Body
\hline 
\multirow{4}{*}{Tasks} & \multirow{4}{*}{Works}  & \multirow{4}{*}{Input} & \multicolumn{4}{c}{Output} & \multirow{4}{*}{Method} & \multicolumn{3}{c|}{Pros (+) and Cons (-)}\\ 
& & & \makecell{Multiple \\ paths} & \makecell{Variable \\ num. of \\paths} & \makecell{Variable \\ path \\ lengths} & \makecell{Long \\ horizon \\ paths} & & \makecell{Fast \\ Inference \\ (+)} & \makecell{Ability to \\ Generalize \\ (+)} & \multicolumn{1}{c|}{For Painting Applications} \\ \hline

\multirow{4}{*}{Spray Painting} & 
\cite{atkar24uniform,Andulkar_Incremental_2015} & \multirow{2}{*}{\makecell[c]{3D \\ (convex only)}}&
  \cmark &
  \xmark &
  \xmark  &
  \cmark &
  \multirow{2}{*}{\makecell[c]{Task-specific \\ Heuristics}} &
  \multirow{2}{*}{\xmark} &
  \multirow{2}{*}{\xmark} & 
  \multirow{2}{*}{\makecell[l]{(+) High paint coverage \\ (-) High design costs and manual tuning}} \\
\cline{2-2} 
\cline{4-7}

  & \cite{Sheng_Automated_2000,Biegelbauer_Inverse_2005,Chen_Automated_2008,Li_Automatic_2010} & & 
  \cmark &
  \cmark &
  \cmark  &
  \cmark & 
  & & & 
  \\ 
  \cline{2-11}

& \cite{Kiemel_PaintRL_2019,jonnarth2024learningcoverageicml} &
  2D &
  \xmark &
  \xmark &
  \xmark  &
  \cmark &
  \makecell{Reinforcement\\ Learning} & 
  \xmark  &
  \xmark &
  \makecell[l]{(+) Explicit paint coverage optimization\\ (-) Requires accurate simulation} \\ \hline

\makecell{Multi-Agent \\ Visual Inspection} & 
\cite{jing2020multi,multiUAV_2023} & 3D &
  \cmark &
  \xmark &
  \cmark  &
  \cmark &
\makecell{Coverage \\ Path Planning} &
  \xmark &
  \xmark & 
  \makecell[l]{(+) High inspection coverage\\ (-) Sample-specific hyperparameters\\ (-) Unable to model painting patterns}
  \\ \hline

\multirow{3}{*}{\makecell{Multi-Agent \\ Motion Prediction}} & \multirow{2}{*}{\cite{yuan2021agentformer,ngiam2022scenetr,varadarajan2022multipath++,nayakanti2023wayformer}} & \multirow{2}{*}{\makecell{2D map \\ + agent features }}  & 
\multirow{2}{*}{\cmark} &
\multirow{2}{*}{\xmark} &
\multirow{2}{*}{\xmark} &
\multirow{2}{*}{\xmark} &
\multirow{7}{*}{\makecell{Imitation and \\ Supervised Learning}} &
\multirow{7}{*}{\cmark} &
\multirow{7}{*}{\cmark} &
\multirow{7}{*}{\makecell[l]{(+) Learns painting patterns from data\\ (+) Little domain knowledge required\\ (-) Implicit paint coverage optimization}} \\ 

& & 
& & & & & & & & \\ \cline{2-7}

& \cite{gu2023vip3d} & 3D &
\cmark &
\cmark &
\xmark & 
\xmark &
 & &&\\ \cline{1-7}

\makecell{3D Shape\\ Completion} &
\cite{Yuan_Pcn_2018,Alliegro_Denoise_2021}
& 3D & \multicolumn{4}{c|}{$\sim$ Point-wise predictions}
&
& &&\\ \cline{1-7}

\makecell{\textbf{Object Centric} \\ \textbf{Motion Generation}} & \textbf{Ours} & 3D &
\cmark &
\cmark &
\cmark & 
\cmark
&
& &&
\\ \hline

\end{NiceTabular}%
}
\end{table*}


\subsection{Robotic Spray Painting}
\label{sec:painting}
Autonomous robotic spray painting is an instance of the NP-hard \emph{Coverage Path Planning} (CPP) problem with additional challenges arising from the non-linear dynamics of paint deposition and hard-to-model engineering requirements. 
Due to its complexity, the landscape of robotic spray painting is dominated by heuristic methods operating under simplifying assumptions about the output path structure (\eg, raster patterns) and object geometry (\eg, convex surfaces)~\cite{Sheng_Automated_2000,Chen_Automated_2008,Li_Automatic_2010,Andulkar_Incremental_2015,atkar24uniform,Chen_Trajectory_2020,gleeson2022generating}.
Most methods further require a 3D mesh or the full CAD model of the object, while point clouds---which are easier to obtain in real-world scenarios through laser scanning---are only considered in~\cite{Chen_Trajectory_2020}.
Critically, all existing heuristics yet assume to work with objects that can be partitioned into convex or low-curvature surfaces. This renders them inapplicable for painting concave objects such as shelves and containers, where global reasoning and more complex path patterns are required. 
Other works rely on matching the objects with a combination of hand-designed elementary geometric components collected in a database~\cite{Biegelbauer_Inverse_2005}.
Matching components are associated with local painting strokes, which are then merged to form painting paths.
Despite its merits, this method requires costly work by experts to explicitly codify object parts and their corresponding painting procedures for each object family and is unable to generalize to arbitrary free-form objects.
Recently, Gleeson et al.~\cite{gleeson2022generating} proposed a trajectory optimization procedure for spray painting that targets the adaptation of an externally provided trajectory candidate, without directly handling motion generation. 

\emph{Reinforcement Learning} (RL) has alternatively been employed to train path generators by directly optimizing objectives such as paint coverage~\cite{Kiemel_PaintRL_2019} or total variation~\cite{jonnarth2024learningcoverageicml}, but these efforts have so far been confined to planar domains.
RL-based stroke sequencing has also shown success in reconstructing 2D images~\cite{Huang_Learning_2019}. Although promising, RL is yet to be demonstrated successful for long-horizon 3D object planning due to the high dimensionality of the state and action spaces. The need for an accurate simulator and low generalizability of RL agents to novel objects also stand out as major issues.

Overall, we remark that all the aforementioned works only show results on a few proprietary object instances. They do not release either the data or the method implementation to allow a fair comparison, besides lacking a discussion on the generalization to new object instances and categories.

Within the CPP literature, robotic applications for multi-agent visual inspection of 3D objects also share important similarities to the OCMG setting.
Recent works proposed optimization-based methods for planning multiple paths and demonstrated their effectiveness in multi-UAV missions on large structures~\cite{jing2020multi,multiUAV_2023}. While these methods can effectively generate long-horizon paths around both convex and concave surfaces, 
they rely on sample-specific hyperparameters, incur high computational costs, and are unable to replicate expert painting patterns.

Table \ref{tab:literature_review} provides a summary of the most relevant publications showing how existing settings and tasks in the literature relate to the OCMG problem.

\section{Method}
\label{sec:method}
\begin{figure}[!t]
    \centering
    \includegraphics[width=\linewidth]{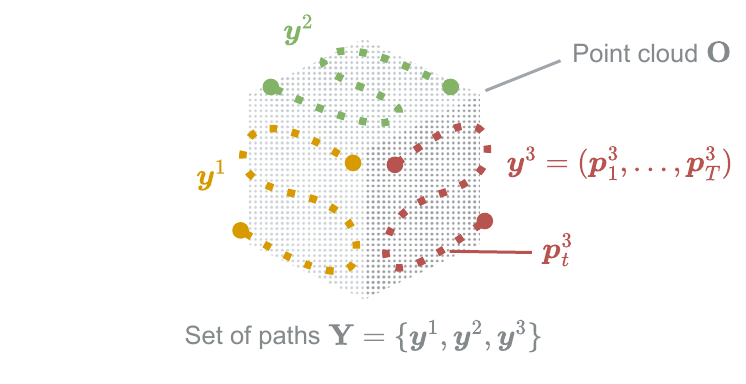}
    \vspace{-16pt}
    \caption{Schematic illustration of a data sample $(\bO, \bY)$ describing input and output of the OCMG problem. The output paths are unstructured as they vary in number and length depending on the input object and can be executed in arbitrary order.}
    \label{fig:notation_helper}
\end{figure}

\subsection{Problem statement}
We formalize the OCMG problem as finding a mapping from a 3D object point cloud to a set of unstructured output paths (see Fig.~\ref{fig:notation_helper}). 
Given expert demonstrations to learn from, we aim to generate accurate paths for previously unseen objects. 

Let $\bO$ represent the object geometry as a point cloud consisting of an arbitrary number of points in 3D space.
Each object $\bO$ is associated with a ground truth set of paths $\bY {=} \{\by^i \}_{i=1}^{n(\bO)}$, with object-dependent cardinality $n(\bO)$. 
Every path $\by^i = (\bp^i_1, \dots, \bp^i_T) \in \mathcal{Y} \subseteq \mathbb{R}^{6 \cdot T}$ is encoded as a sequence of 6D poses $\bp \in \mathbb{R}^6$. For simplicity, we consider the dimension $T$ fixed, with shorter paths zero-padded to reach that maximum length.  

Under this formulation, we consider the problem of finding a function $f : 2^{\mathbb{R}^{3}} \rightarrow 2^{\mathcal{Y}}$ mapping the set of points $\bO$ describing the object geometry to the set of desired paths $\bY$\footnote{For a set \( \mathcal{S} \), the notation \( 2^{\mathcal{S}} \) here denotes the powerset of \( \mathcal{S} \), \ie, the set of all subsets of \( \mathcal{S} \).}.
To do so, we parametrize $f$ using a deep neural network, and train it through empirical risk minimization~\cite{vapnik1991principleserm}.
Specifically, we minimize a loss function $\mathcal{L}(\hat{\bY}, \bY)$, which quantifies the discrepancy between the predicted paths $\hat{\bY}=f(\bO)$ and the ground truth paths $\bY$ on the training data, using gradient descent to optimize the network parameters.

We highlight that this formulation does not make task-specific assumptions related to the spray painting problem, making our contribution applicable to a broad range of object-centric motion generation tasks (\eg, welding or cleaning).

\subsection{Method Overview}
We tackle object-centric motion generation with a tailored deep learning model that copes with unstructured input---3D point clouds---and unstructured output paths.
More precisely, instead of directly predicting a set of paths, we decompose the problem into the prediction of unordered path segments, \ie, short sequences of 6D end-effector poses (or \emph{waypoints}).
In addition, we concurrently predict a set of probability masks that identify which segments belong to the same path.
We denote our method as \emph{\ours}, and emphasize that all required path segments and masks are predicted by our network in parallel, with a single forward pass.
Such approach induces our model to learn end-to-end global embeddings of the input object that allow for both local (segments) and global (masks) planning decisions in one step.
Overall, by designing a joint segment and mask prediction pipeline, we conveniently simplify the problem of dealing with unstructured paths and address 

\begin{figure}[!t]
    \centering
    \includegraphics[width=\linewidth]{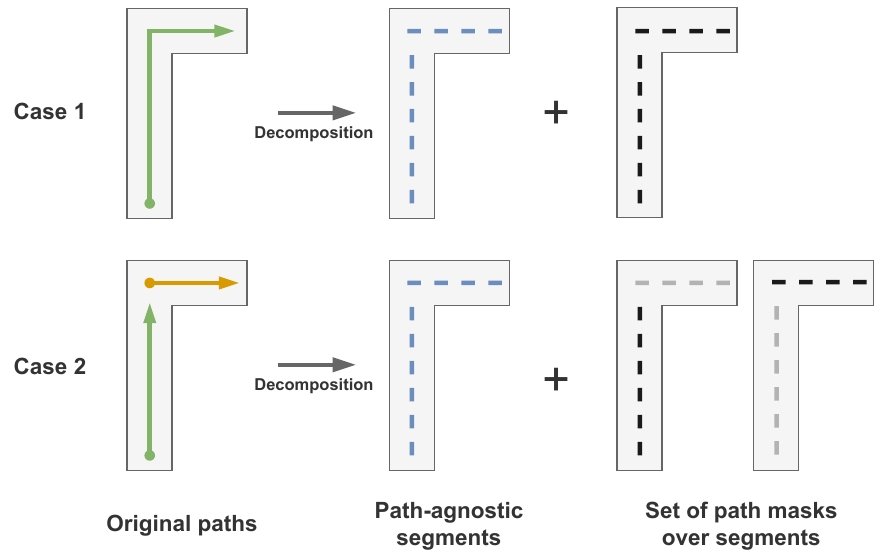}
    \vspace{-8pt}
    \caption{Example of two arbitrary configurations of ground truth paths $\bY$ on an L-shaped 2D object.
    Notice how \ours can easily manage both cases by breaking down the learning problem into the prediction of path-agnostic segments and their associated path masks.
    }
    \label{fig:temporal_correlation}
    \vspace{-6pt}
\end{figure}

\begin{itemize}[leftmargin=*,itemsep=2pt]
\item \textbf{long-horizon paths:} we do not impose constraints on the shape or length of each path.
Diverse path configurations can be handled just as effectively, as exemplified in Fig.~\ref{fig:temporal_correlation}.
\item \textbf{unordered paths:} within a set prediction framework, the order of the predicted path segments is irrelevant. Therefore, concatenating segments that belong to the same path naturally yields a set of output paths that are invariant by permutation.
\item \textbf{variable length and number of paths:} we predict a conservatively large number of path segments, allowing for the generation of redundant overlapping segments that can be easily filtered out.
\end{itemize}

Our method takes inspiration from the Panoptic Segmentation (PanSeg) literature~\cite{carion2020detr,cheng2021maskformer}, where a variable number of class instances shall be predicted given a static environment with global information (an RGB image).
Notably, works in the field of PanSeg shifted towards one-shot predictors over the years as opposed to multiple-stage or autoregressive approaches.
Similarly, we depart from sequential methods for OCMG, as we aim to reach real-time inference capabilities and avoid compounding errors on long-horizon predictions.

Throughout this work, we demonstrate that \ours is capable of predicting a large number of long-horizon paths with a single forward pass, that implicitly incorporate task-specific requirements without domain knowledge.
We remark that ad-hoc trajectory optimization methods~\cite{gleeson2022generating} may still be applied to impose task-specific kinodynamic constraints (\eg  reachability and collision avoidance), which we consider subsequent and complementary to the scope of this work. 
Here, we directly interpolate the predicted sequence of 6D waypoints for execution on a real robot, resulting in feasible trajectories out of the box.

In the following part of this section we describe each step of our method in detail, breaking it down into segment predictions (Sec.~\ref{sec:segments_prediction}), mask predictions (Sec.~\ref{sec:masks_prediction}), and postprocessing (Sec.~\ref{sec:postprocessing}).

\begin{figure*}[!t]
    \centering
    \includegraphics[width=\linewidth]{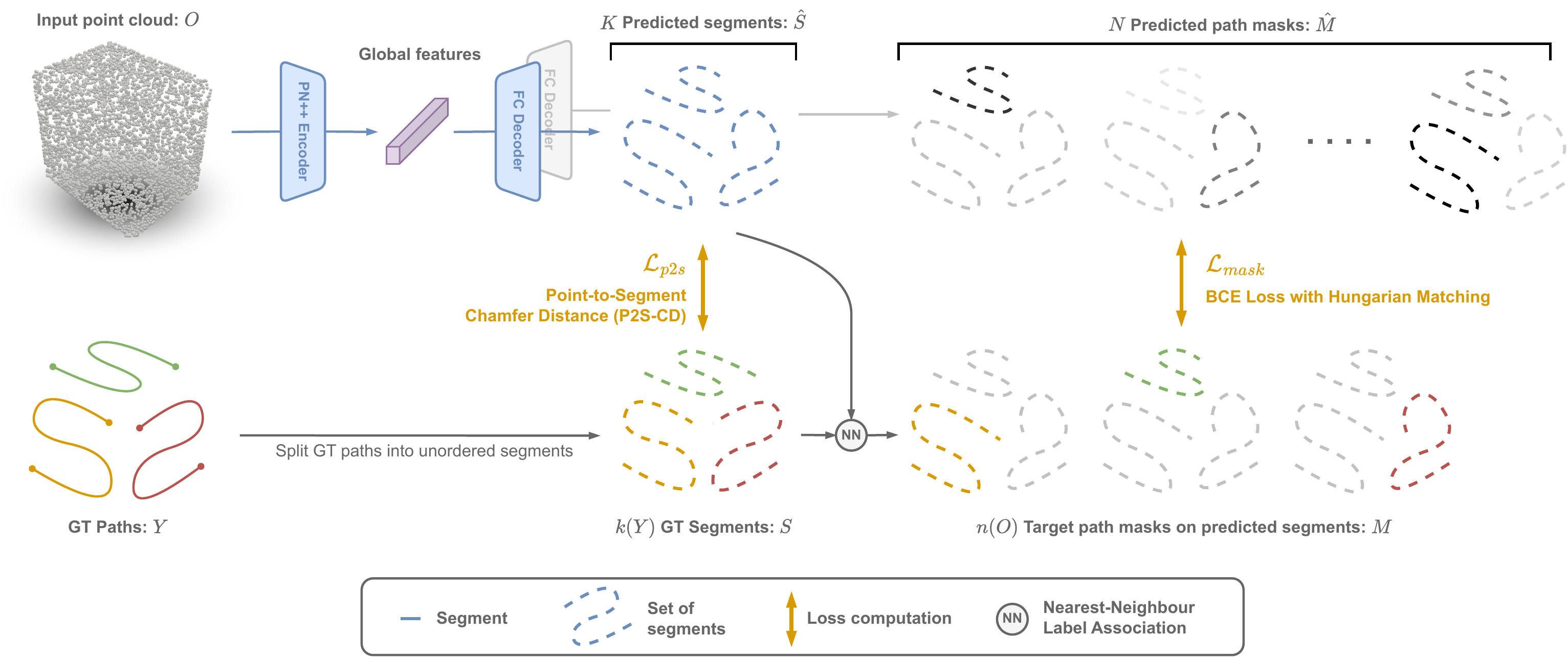}
    \caption{Overview of the training pipeline of our method (\ours). Global features are learned from a point cloud representation of the input object, and used to concurrently predict path segments and path masks, in a single forward pass.
    }
    \label{fig:method}
    \vspace{-6pt}
\end{figure*}


\subsection{Segment Predictions}
\label{sec:segments_prediction}

Let $\bS$ be the set of ground truth path segments $\bS {=} \{\bs^j\}_{j=1}^{k(\bY)}$ of an object $\bO$ and its associated paths $\bY$.
We define a segment as a sequence of $\lambda \in \mathbb{N}^+$ poses obtained from a path $\by^i \in \bY$.  
Namely, $\bs^j = (\bp^i_t, \dots, \bp^i_{t+\lambda-1}) \in \mathbb{R}^{6 \cdot\lambda}$ for some timestep $t{=}1, \dots , T-\lambda$.
We derive all segments in \(\bS\) by striding along all paths with a step of $\lambda-1$, \ie considering an overlap of one pose between consecutive segments.
Notice that the total number of resulting segments $k(\bY)$ depends on the number of paths and their lengths.

We design our model to take the object point cloud $\bO$ as input, and predict a set of path segments $\bShat{=}\{\bshat^j\}_{j=1}^{K}$ that approximate the true segments $\bS$. 
We adopt the PointNet++ architecture~\cite{Qi_Pointnet++_2017} as basic backbone for global feature extraction from $\bO$, followed by a fully connected 3-layer decoder that jointly outputs all path segments. 
A fixed number of segments $K = \max k(\bY)$ is predicted to ensure all ground truth segments are recovered for all objects.

Let $\bP$ be the set of unordered ground truth waypoints, 
formally described as 
$\{\bp^i_t {\in} \mathbb{R}^6 \ | \ i {\in} [1, \dots, n(\bO)], \ t{\in} [1, \dots, T] \}$
---equivalent to the set of segments for $\lambda{=}1$. Analogously, we define  
$\bPhat{=}\{ \bshat^j_t {\in} \mathbb{R}^6 \ | \ j {\in} [1, \dots, K], \ t {\in} [1, \dots, \lambda] \}$
as the set of individual predicted waypoints, obtained by interpreting $\bShat$ as an unordered collection of waypoints.

\ours is trained with a novel loss function aimed at driving the prediction of path segments $\bShat$ close to the ground truth segments $\bS$ by means of Euclidean distances in $\mathbb{R}^{6\cdot\lambda}$ space.
To do this, our loss includes auxiliary point-wise terms that penalize prediction errors in the lower dimensional space $\mathbb{R}^{6}$, by disregarding how waypoints are arranged into segments.
Overall, our \emph{Point-to-Segment Chamfer Distance} (P2S-CD) is defined as:
\begin{equation}
\label{eq:pscd}
\begin{split}
    \mathcal{L}_{p2s}(\bPhat ,\bP,\bShat ,\bS) & = \\
    & w^{f}_{p} \cdot d_{ACD}(\bPhat , \bP) + w^{f}_{s} \cdot d_{ACD}(\bShat , \bS) \ + \\
    & \underbrace{\raisebox{10pt}{$w^{b}_{p} \cdot d_{ACD}( \bP,\bPhat)$}}_{\text{Point-wise}} \raisebox{10pt}{$+$} \underbrace{\raisebox{10pt}{$w^{b}_{s} \cdot d_{ACD}(\bS,\bShat)$~.}}_{\text{Segment-wise}}
\end{split}
\end{equation} 
Here, the \emph{Asymmetric Chamfer Distance} (ACD) \emph{from} set $A$ \emph{to} set $B$ is: 
\begin{equation}
    d_{ACD}(A, B) = \frac{1}{|A|} \sum _{a\in A}\min_{b\in B} \| a-b\| _{2}^{2}~,
\end{equation}
and the parameters $w^f_p, w^f_s, w^b_p, w^b_s \in \mathbb{R}^{+}$ weight the computation of the four ACD terms ($f$: forward, $b$: backward, $p$: point-wise, $s$: segment-wise). 

\begin{figure}[tb]
    \centering
    \includegraphics[width=\linewidth]{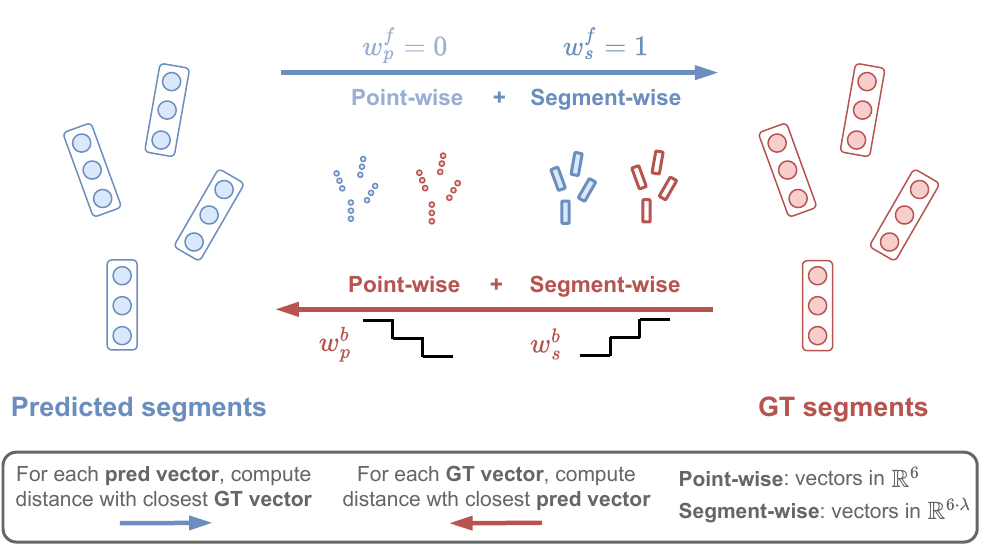}
    \caption{
    Illustration of our Asymmetric Point-to-Segment curriculum for segment predictions ($\lambda{=}3$). The parameters $w^b_p,w^b_s$ weighting the backward point-wise and segment-wise ACD terms vary during training.}
    \vspace{-6pt }
    \label{fig:apscurriculum}
\end{figure}

Based on this formulation, we propose an auxiliary point-to-segment curriculum that (1) first focuses on matching waypoints in some random permutation (point-wise terms), and then (2) gradually promotes local structure by computing distances among sequences of poses (segment-wise terms).
To achieve this, we start the training with a dominant point-wise term ($w^b_p {\gg} w^b_s$) and progressively converge to equal point-wise and segment-wise contributions ($w^b_p{=}1, w^b_s{=}1$).
Notably, this procedure is asymmetric and only applied on the backward terms: this ensures that predictions are pulled sufficiently close to the ground truth poses, which in turn enables effective global coverage of the desired poses, and limits the generation of clusters of poses~\cite{densityawarecd}.
On the other hand, we keep the forward weights fixed to $w^f_p{=}0$, $w^f_s{=}1$ throughout the training, promoting the generation of segments that are smooth and locally accurate.
We refer to this process as the \emph{Asymmetric Point-to-Segment} (AP2S) curriculum, and report a schematic illustration of our overall loss function in Fig.~\ref{fig:apscurriculum}.

\begin{figure*}[!t]
    \centering
    \includegraphics[width=\linewidth]{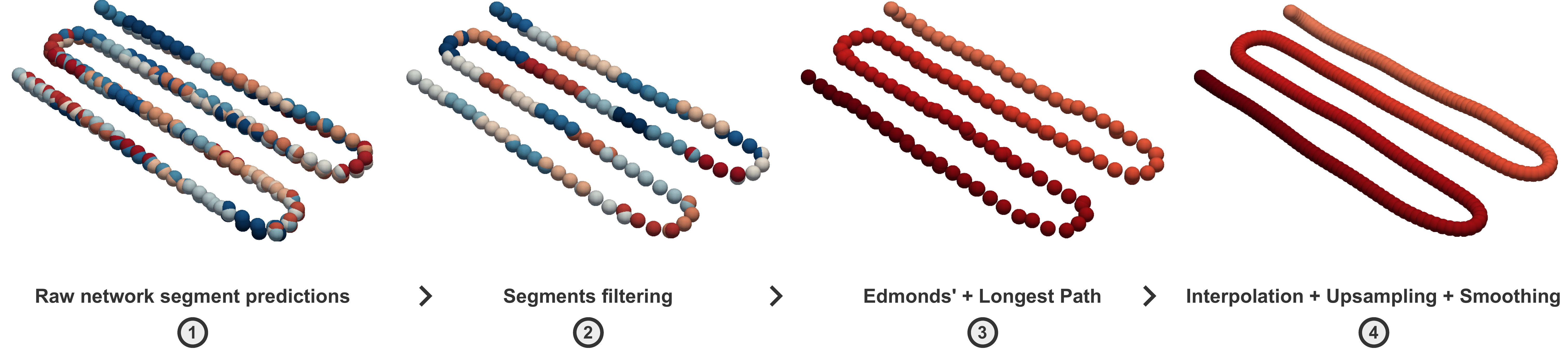}
    \caption{
    Postprocessing: concatenation of the set of predicted segments belonging to the same path mask.
    In step (1) the figure depicts raw network predictions with $\lambda{=}4$, with separate segments differentiated by color. Step (2) shows the effect of segment filtering. In step (3) and (4), where the path is identified and further refined, the ordered sequence of waypoints is shown with a color gradient.}
    \label{fig:postprocessing}
\end{figure*}

\subsection{Mask Predictions}
\label{sec:masks_prediction}

Our model concurrently predicts a set of probability masks over predicted segments, indicating which segments belong to the same path. 
Defining a supervised learning objective for this task is not trivial, as it requires comparing predicted masks on generated segments with ground truth masks on generated segments. However, the latter do not exist.
A reasonable choice is to match each predicted segment with the closest ground truth segment---as already proposed in Sec.\ref{sec:segments_prediction}---and construct target masks accordingly.
We refer to this strategy as \emph{nearest-neighbour label association}, and use it to project masks over ground truth segments onto predicted segments.
In particular, let $\bM{=}\{\bm^i\}_{i=1}^{n(\bO)}$ be the set of \emph{target path masks} for some input pair $(\bO,\bY)$.
Each element $\bm^i \in \{0,1\}^K$ encodes the set of predicted segments that belong to path $i$, that is
\begin{equation}
\begin{split}
    & \bm^{i}_{j} =\begin{cases}
1 & \mathrm{if\ } \mathrm{NN}(\bshat^j) \ \mathrm{belongs \ to \ path} \ i\\
0 & \mathrm{otherwise},
\end{cases} \\
& \forall \ j = 1, \dots, K
\end{split}
\end{equation}
where $\mathrm{NN}(\bshat) {=} \argmin_{\bs \in \bS} \| \bs-\bshat \|^2$. 
Our model is designed to predict a set of $N\geq n(\bO)$ probability masks $\bMhat{=}\{ \bmhat^{i} \}_{i=1}^{N}$, where $N{=}\max n(\bO)$ is the maximum number of paths across objects $\bO$ in the training set. 
All masks are predicted in parallel via a 3-layer MLP decoder from the global features of the object, followed by a sigmoid activation.
Note that, as in~\cite{cheng2021maskformer}, we do not force mask predictions to be mutually exclusive for a given segment $\bshat^j$. Hence we avoid using a softmax activation.

We drive the predicted path masks $\bMhat$ towards the target path masks $\bM$ through the Binary Cross Entropy (BCE) loss:
\begin{equation}
\label{eq:bce_loss}
    \mathcal{L}_{bce}(\bmhat, \bm) = - \sum_{j=1}^{K}\Bigl[\bm_j\cdot\log(\bmhat_j) + (1-\bm_j)\cdot\log(1-\bmhat_j)\Bigl].
\end{equation}
Importantly, we must consider the prediction of permutation invariant masks to cope with a set of unordered paths.
Therefore, we assign each predicted path mask $\bmhat$ to a target mask $\bm$ by finding a bijection $ \sigma : \bMhat \rightarrow \bM$.
Similarly to~\cite{carion2020detr,cheng2021maskformer}, we do this by solving a bipartite matching problem between the two sets, where the assignment costs are computed using $\mathcal{L}_{bce}$.
Particularly, we pad the target masks with a ``no path'' token $\varnothing$ to allow one-to-one matching.
Ultimately, the training loss for mask prediction is as follows:
\begin{equation}
\begin{split}
    \mathcal{L}_{mask}(\bMhat, \bM) = \\ \sum_{i=1}^{N}
    \Bigl[ 
        & c^{\sigma(i)}\cdot\log(\hat{c}^i) + (1-c^{\sigma(i)})\cdot\log(1-\hat{c}^i) \\
        & + c^{\sigma(i)} \cdot
         \mathcal{L}_{bce}(\bmhat^i, \bm^{\sigma(i)})
    \Bigl]~,
\end{split}
\end{equation}
where $\hat{c}^i \in [0,1]$ is a learned confidence score for each mask, and $c^i{=}\mathds{1}_{\bm^i \neq \varnothing}$ indicates the true masks.
Overall, we train our model to minimize $\mL = \mL_{p2s} + \mL_{mask}$. 
We display a schematic description of our training pipeline in Fig.~\ref{fig:method}.

At inference time, we derive the groups of segments belonging to the same path by assigning each predicted segment $\bshat^j$ to one of the $N$ predicted masks. Formally, the assignment of segment $\bshat^j$ occurs as follows:
\begin{equation}
\label{eq:path_mask_assignment}
    \begin{split}
        \underset{i \in [1, \dots, N]}{\argmax} \ \bmhat^i_j \ \ \text{s.t.} \ \ \hat{c}^i \geq 0.5~.
    \end{split}
\end{equation}
In other words, segments are assigned to the path mask with the highest predicted probability, discarding path masks that are predicted as ``no path''. Assigning all predicted segments according to Eq.~(\ref{eq:path_mask_assignment}) results in a final number of $\hat{n}(\bO)\leq N$ paths predicted by our model given the input object, ideally equal to $n(\bO)$.

\subsection{Postprocessing: Segment Concatenation}
\label{sec:postprocessing}
A final postprocessing step is applied to concatenate the subset of predicted segments $\bShat^i \subseteq \bShat$ that are assigned to the same path mask $\bmhat^i$, and produce an ordered sequence of 6D waypoints, \ie the executable path.

Viable solutions may include solving an open Traveling Salesman Problem (TSP) among segments or employing learning-based approaches for ranking. Here, we adopt a simple and effective concatenation strategy based on segment proximity and alignment. 

First, predicted segments in excess are removed by discarding segment pairs whose distance falls below a predefined threshold, proceeding in ascending order of pairwise distances\footnote{Note that our loss $\mathcal{L}_{p2s}$ implicitly promotes overlapping if the number of predicted segments $K$ is higher than the ground truth segments $k(\bY)$.}. 
Then, we find an optimal path that connects the retained segments. Consider the set of segments in $\bShat^i$ as nodes of a directed graph. An edge among two segments is weighted based on the proximity in space and orientation between the starting and ending poses of the two segments, as well as the similarity in segment directions.
More formally, the assigned cost to the edge from $\bshat^j$ to $\bshat^k$ is 
\begin{equation}
\label{eq:cost}
    C(\bshat^j,\bshat^k) = \| \bshat^j_\lambda - \bshat^k_1 \|_2^2 + w_v \cdot \| (\bshat^j_\lambda - \bshat^j_{\lambda-1}) - (\bshat^k_2 - \bshat^k_1)  \|_2^2~.
\end{equation}
Here, $w_v \in \mathbb{R}^+$ is a trade-off weight between the two terms, and the segments' subscripts specify the index of a particular pose among the $\lambda$ poses that make up each segment.

We find the optimal concatenation by employing the Edmonds' algorithm~\cite{edmonds1967optimum} to the $k$-nearest neighbor graph of segments $\bShat^i$ constructed using $C(\cdot, \cdot)$ and $k{=}5$.
Ultimately, we extract the longest path from the resulting Directed Acyclic Graph (DAG) and obtain the final \emph{ordered} sequence of predicted segments.
At this point, filtering techniques such as interpolation, upsampling, and smoothing may be conveniently applied.
The same process is repeated independently for all predicted path masks.
For clarity, we illustrate each step of the postprocessing in Fig.~\ref{fig:postprocessing}.

\section{The Extended PaintNet Dataset}
\label{sec:ext_paintnet_dataset}
\begin{figure*}[!t]
    \centering
    \includegraphics[width=\linewidth]{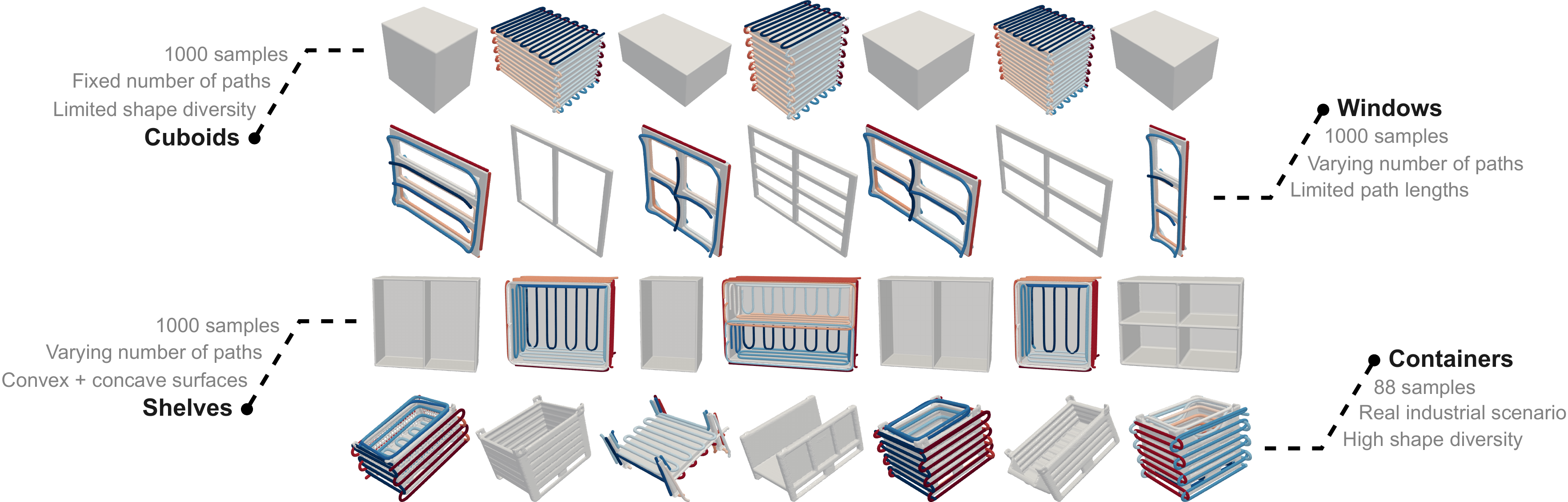}
    \caption{Overview of a number of representative instances of the Extended PaintNet dataset, featuring realistic spray painting demonstrations designed by experts. The dataset is divided into four categories of growing complexity.
    Different colors represent separate paths.
    }
    \label{fig:dataset_overview}
    \vspace{-6pt}
\end{figure*}

PaintNet~\cite{tiboni2023paintnet} was the first dataset of expert demonstrations introduced to support the study of motion generation conditioned on free-form 3D objects, and was specifically designed for the spray painting task. In this work, we expand PaintNet more than threefold, resulting in a new version that contains 3088 samples.

Every sample is a pair of a 3D object and its corresponding spray painting paths. Each object is represented as a triangle mesh, with vertex coordinates expressed in real-world millimeter scale. The meshes are provided in an aligned and smoothed watertight~\cite{Huang_Robust_2018} format with any private information (\eg, engraved logos) accurately anonymized.
The number of spray painting paths associated to an object varies according to the geometry of the object. Each path is encoded as a sequence of end-effector configurations in task space, \ie, positions and orientations of the gun nozzle. More precisely, we record 3D positions as the ideal paint deposition point, 12cm away from the gun nozzle, and 3D gun orientations as Euler angles.
The data pairs represent realistic spray painting demonstrations that a real robot could directly execute. In other words, they encode feasible trajectories designed by experts to reach near-complete coverage of the whole surface of the 3D objects. Each waypoint is collected by sampling the end-effector pose at a rate of 250Hz during execution. 
Ground-truth paths are produced ad-hoc for each object category by custom heuristics based on long-standing experience in the field.

The four object categories composing the dataset are presented below, ordered by increasing complexity:

\begin{itemize}[leftmargin=*,itemsep=2pt]
 \item \textbf{Cuboids}: a basic class of 1000 rectangular cuboids tailored for testing models under minimal generalization requirements and relatively simple path patterns. Cuboids are sampled with varying height and depth uniformly from $0.5m$ to $1.5m$, while having a fixed width of $1m$. Their volume ranges from $0.25m^3$ to $2.25m^3$. A fixed number of six raster-like paths is associated to each cuboid to paint the exterior faces, with gun orientations that are perpendicular to the surface at all times.

\item \textbf{Windows}: a set of 1000 window-like data pairs, with width and height varying uniformly from $0.4\,m$ to $1.8\,m$, and a fixed thickness of $4\,cm$. For each sample, up to $3$ horizontal cross sections and up to $1$ vertical cross section are randomly selected. Windows introduce harder challenges for motion generation, such as predicting a variable, high number of paths while handling non-trivial gun orientations and geometric patterns.

\item \textbf{Shelves}: a set of 1000 shelves featuring highly concave surfaces. The instances differ significantly in volume and number of inner compartments. Their volume ranges from $18dm^3$ to $160dm^3$, with up to 6 inner compartments for the larger samples.

\item \textbf{Containers}: a set of 88 industrial containers including meshes with highly heterogeneous global and local geometric properties (\eg, wavy and grated surfaces). Here, the manually-guided painting paths designed by experts show evident irregularities across instances. This class of objects is particularly challenging due to the limited number of samples.
\end{itemize}
An overview of the Extended PaintNet dataset is depicted in Fig.~\ref{fig:dataset_overview}.
The dataset is publicly available at \url{https://gabrieletiboni.github.io/MaskPlanner/}.

\section{Experimental Evaluation}
\label{sec:experiments}
\subsection{Implementation Details}
\label{sec:implementation_details}
\noindent \textbf{Data preparation.}
All experiments are carried out on the Extended PaintNet dataset introduced in Sec.~\ref{sec:ext_paintnet_dataset}.
Input point clouds $\bO$ are derived by sampling 5120 points from the surface of the available meshes through Poisson Disk sampling~\cite{Cook_Stochastic_1986}.
We further down-sample ground-truth paths so that adjacent poses are approximately 5cm apart, avoiding to deal with needlessly dense waypoints.
Each waypoint is represented as a 6D vector with position described by $(x,y,z)$ coordinates, and orientation encoded as a 3D unit vector, rather than Euler angles. This simplification is permitted by our conic spray gun model, which is symmetric and invariant to rotations around the approach axis.
Consequently, ground-truth Euler angles are converted into 3D unit vectors (2-DoF), indicating where the gun nozzle is pointing.
When computing Euclidean distances in Eq.~(\ref{eq:pscd}) we rescale and adjust the relative importance of position and orientation components in the 6D vector by weighting the latter by $0.25$. 
Point clouds and path poses in the dataset are finally transformed so that each sample is centered around the origin, and down-scaled by a dataset-global factor.
We create training and test sets by dividing each category in the dataset according to an 80\%/20\% split. 

\begin{figure*}[!t]
    \centering
    \includegraphics[width=0.75\linewidth]{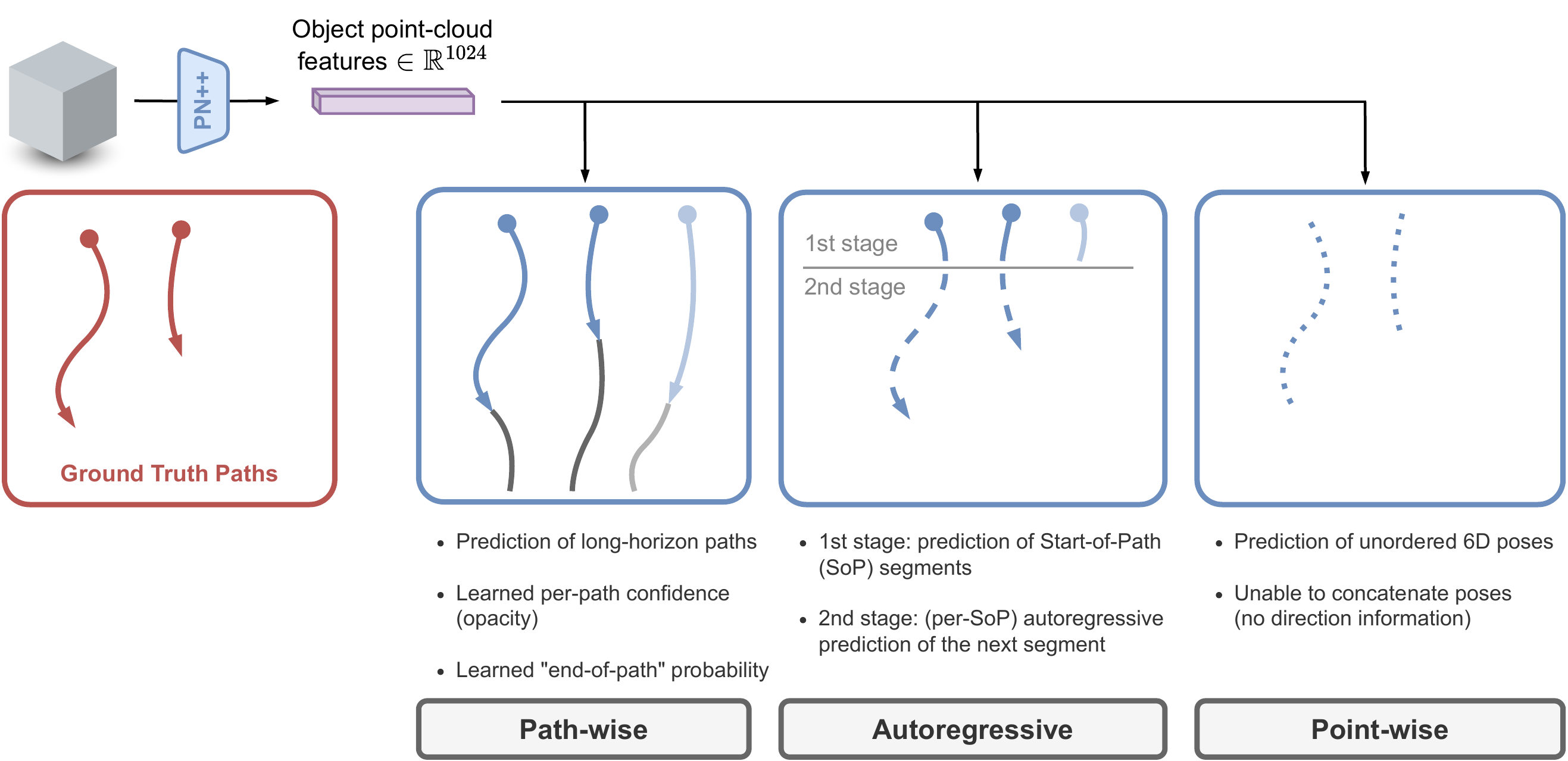}
    \caption{Overview of the novel baselines implemented for comparison with \ours in the OCMG problem. All baselines share the same PointNet++ (PN++) encoder architecture for extracting object features.
    }
    \label{fig:baselines_overview}
    \vspace{-6pt}
\end{figure*}

\noindent \textbf{Architecture.} We employ an encoder module based on PointNet++~\cite{Qi_Pointnet++_2017} as the feature extractor mapping the input point cloud to a latent space of dimensionality 1024.
The encoder is initialized with pre-trained weights from an auxiliary shape classification task on ModelNet~\cite{wu_modelnet_2014}.
Segment and mask predictions are generated in parallel from the object features through separate dedicated 3-layer MLPs decoders with hidden sizes (1024,1024) and output sizes $6{\times}\lambda{\times} K$ and $K{\times} N$---respectively for segments and masks.
Confidence scores $\hat{c}^i$ for each path mask are learned through an additional linear output layer of size $N$, followed by a sigmoid activation.
In line with~\cite{tiboni2023paintnet}, we set $\lambda{=}4$ throughout all our experiments.

\noindent \textbf{Training details.}
We minimize our loss function $\mL$ using the Adam optimizer with learning rate $10^{-3}$ over $4800$ epochs.
Notably, we find it beneficial to activate the path mask loss $\mL_{mask}$ only after $3200$ epochs, so that masks are learned only when the predicted segments are sufficiently close to the ground truth.
We implement the AP2S curriculum described in Sec.~\ref{sec:segments_prediction} by scaling $w^b_p$ and $w^b_s$ by factors of $0.1$ and $10$, respectively, at epochs 1000 and 2000.
The learning rate is halved five times at intervals of 800 epochs throughout the training process.
Training takes approximately $6$ hours on a single NVIDIA RTX-4080 GPU for a training set of $800$ objects and a batch size of $64$.
Each forward pass on 5120 input points takes approximately 100ms on the same GPU, while postprocessing requires 100-500ms on a single CPU core, depending on the number and length of output paths.

\subsection{Baselines}
The OCMG problem as described in Sec.~\ref{sec:introduction} and formalized in Sec.~\ref{sec:method}, is introduced in this work for the first time. The overview of previous literature in Sec.~\ref{sec:related_work} highlighted that none of the existing methods are suitable for tackling OCMG tasks via Deep Learning. To provide a thorough benchmark comparison of \ours against alternative approaches, we devise a new set of baselines as ad-hoc adaptations of existing works designed for different problem settings.
A schematic overview of all baselines is depicted in Fig.~\ref{fig:baselines_overview}.
We provide a detailed description of each baseline below.

\noindent\textbf{Path-wise.}
This model outputs a set of complete paths at once, where each path is a $6 {\cdot} T$-dimensional vector.
It predicts a predefined maximum number of paths, and identifies which paths to retain at inference time through learned confidence scores.
The model also predicts an ``end-of-path'' probability for each individual path waypoint, thus the length of each path can vary.
This approach is inspired by the object detection literature, with paths treated analogously to bounding boxes. In particular, we adopt the same logic of one-shot set prediction as in DETR~\cite{carion2020detr}, but we design the network architecture with MLP layers as in \ours, instead of DETR's query-based modules.
Overall, this implementation extends the \emph{Multi-Path Regression} baseline introduced in~\cite{tiboni2023paintnet}, and reflects the attempt to directly predict long-horizon paths as opposed to breaking down the problem into segment predictions.

\noindent \textbf{Autoregressive.}
This baseline allows comparing one-shot prediction methods---such as \ours and Path-wise---to an alternative autoregressive strategy for OCMG.
We design a method for predicting the next pose and a termination probability given a history of previously generated poses, separately for each path. 
It consists of a first model trained to predict a set of Start of Path (SoP) poses given the object features (first stage). This is followed by the autoregressive model conditioned on the predicted SoP, the object features, and the $10$ most recent predictions (second stage).
The training procedure for the first stage is analogous to the Path-wise baseline, \ie, the SoP model learns confidence probabilities for filtering out SoPs in excess.
We then use teacher forcing~\cite{teacherforcing} to train the autoregressive model and inject noise into the input history to mitigate compounding errors.
Interestingly, we observe that predicting a sequence of $\lambda$ output poses---\ie, segments---at each autoregressive step yields smoother output paths.
Thus, we adopt this approach for this baseline.
Finally, we find that jointly training the SoP and autoregressive models at the same time
leads to training instabilities and does not improve predictive performance.

\noindent \textbf{Point-wise.}
Inspired by shape completion methods~\cite{Yuan_Pcn_2018}, the approach proposed in \cite{tiboni2023paintnet} directly predicts all output waypoints as a set of unordered 6D poses, employing the standard symmetric Chamfer Distance~\cite{Fan_Point_2017}.
We integrate this approach with the mask predictions pipeline of \ours in order to recognize how output waypoints are organized into separate paths.
Notably, this implementation can be seen as an edge case of \ours with fully point-wise loss terms $w^f_p{=}1,w^f_s{=}0,w^b_p{=}1,w^b_s{=}0$.
However, unlike \ours, this baseline is unable to concatenate output poses into long-horizon paths through postprocessing, rendering it practically inapplicable to a real-world scenario.

\subsection{Evaluation Metrics}
\label{sec:eval_metrics}
We introduce a collection of metrics for OCMG to assess the performance of different methods in predicting unstructured paths. In the following we use $h{=}1,\ldots,H$ as an index on the test object samples $\bO_h$. 

\noindent \textbf{Point-wise Chamfer Distance (PCD)~\cite{Fan_Point_2017}.} It compares the predicted and ground-truth paths as two clouds of 6D poses by computing $\frac{1}{H}\sum_{h=1}^H PCD(\bPhat_h, \bP_h)$, where $PCD(\bPhat, \bP) {=} d_{ACD}(\bPhat ,\bP) + d_{ACD}( \bP,\bPhat)$. This metric accounts for the accuracy of all end-effector positions and orientations, while disregarding the order among poses. Lower is better.

\noindent \textbf{Accuracy of Number of Paths (Acc-NoP).} It measures the fraction
of objects across the test set for which the predicted number of output paths matches the true number, \ie, $\frac{1}{H}\sum_{h=1}^H\mathbbm{1}_{\hat{n}(\bO_h)=n(\bO_h)}$. Higher is better.

\noindent \textbf{Mean Absolute Error of Number of Paths (MAE-NoP).} This metric measures the average deviation of the predicted number of paths from the true number. It is obtained by 
computing the mean absolute error $\frac{1}{H}\sum_{h=1}^H\vert \hat{n}(\bO_h) - n(\bO_h)\vert$ 
across objects in the test set\footnote{Both MAE-NoP and Acc-NoP are based on established metrics for ordinal regression problems in machine learning (\eg, see Sec.~4.1.3 in~\cite{7161338}).}. 
Lower is better.

\noindent \textbf{Paint Coverage (PC).}
Although not explicitly optimized at training time, we aim to assess the percentage of object surface covered by the predicted paths when executed in a spray painting simulator.
To do so, we follow the same approach as in~\cite{tiboni2023paintnet}.
First, the $10_{th}$ percentile of the ground truth paint thickness distribution across mesh faces is selected as the relative paint thickness threshold for measuring coverage. 
Next, we execute the predicted paths and evaluate the percentage of faces whose paint thickness is larger than such threshold.
We then average the percentages across all instances of the test set. 
Note that the relative thickness threshold makes the metric independent of the specific spray gun model parameters used during simulation (\eg, paint flux), thus renders it suitable for benchmarking purposes.

\begin{figure*}[!t]
    \centering
    \includegraphics[width=\linewidth]{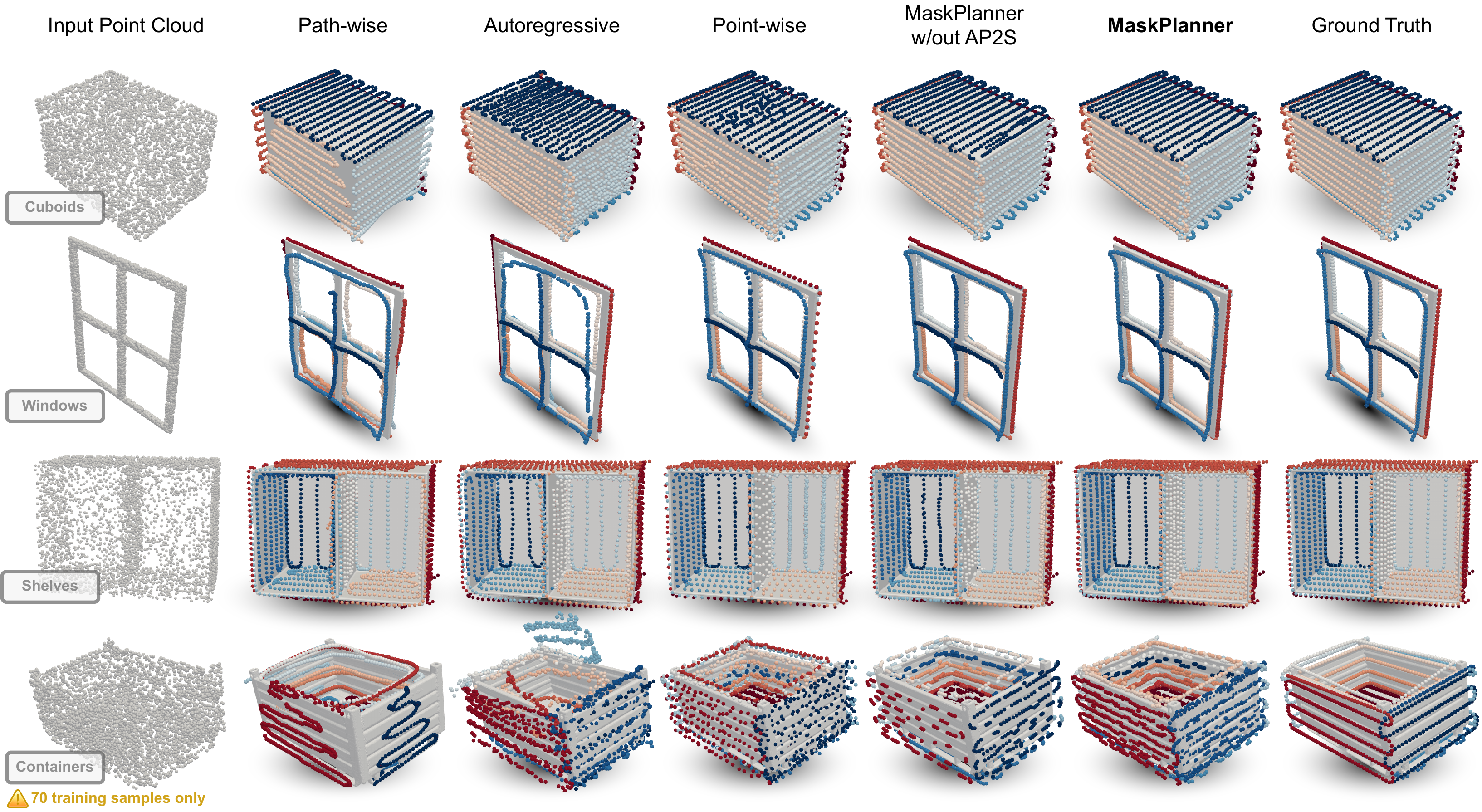}
    \caption{
        Main qualitative results: the raw network predictions of all baselines are shown for a representative test sample of each object Category. Points displayed with the same color belong to the same path. Point orientations are not visible.
    }
    \label{fig:qualitatives_segments_prediction}
\end{figure*}

\begin{figure*}[!t]
    \centering
    \includegraphics[width=\linewidth]{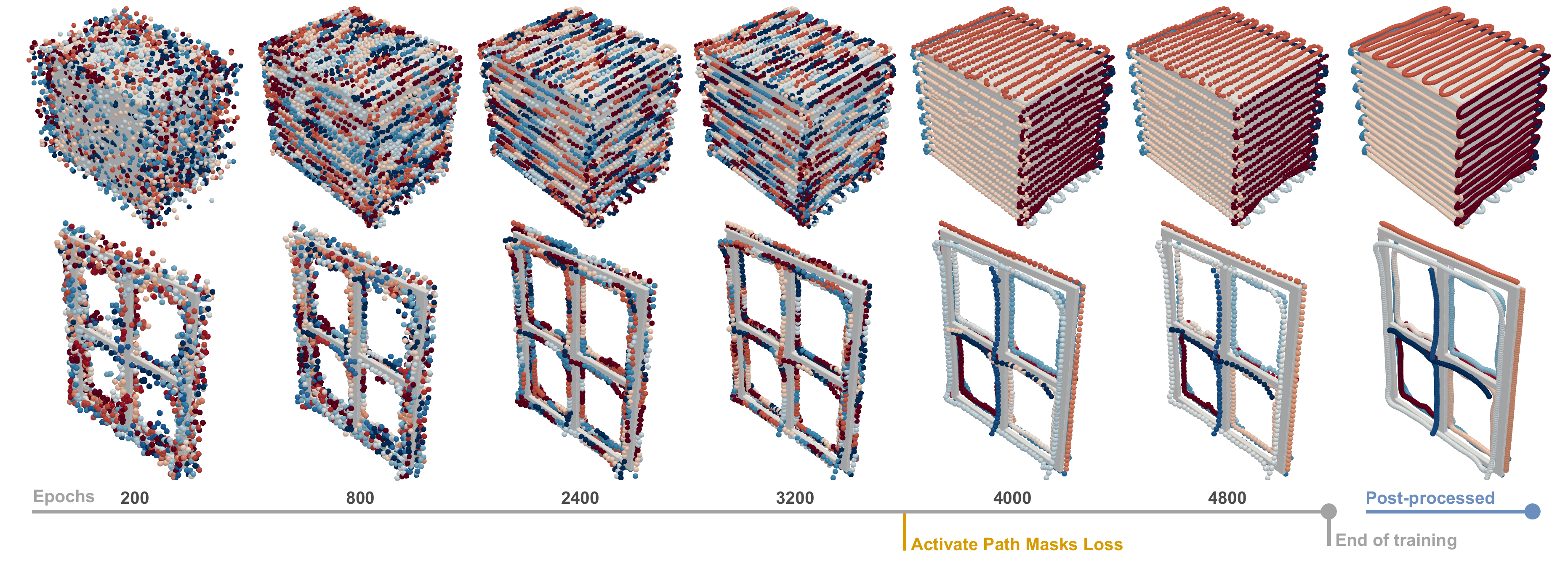}
    \caption{
    The network is trained in a coarse-to-fine manner via the Asymmetric Point-to-Segment curriculum:
    first, output waypoints are positioned across the surface; then, local structure is promoted.
    Segments associated to the same path mask are shown with the same color.
    }
    \label{fig:intermediate_results}
\end{figure*}

\begin{table*}[]
\centering
\caption{
Main quantitative results on the test set of each Category. We report Mean \( \pm \) St.dev for 10 training repetitions.
Mean values that are \emph{not} statistically significantly worse than any other are marked in bold ($\alpha {=} 0.05$). Remind that all methods predict a fixed number of six output paths for the Cuboids category, hence no NoP metrics are computed.
}
\label{tab:quantitative_all_metrics}
\resizebox{\textwidth}{!}{%
\def\arraystretch{1.25}%
\begin{tabular}{l|c|ccc|ccc|ccc|}
\cline{2-11}
 &
  Cuboids &
  \multicolumn{3}{c|}{Windows} &
  \multicolumn{3}{c|}{Shelves} &
  \multicolumn{3}{c|}{Containers} \\ \cline{2-11} 
 &
  \begin{tabular}[c]{@{}c@{}}PCD\\ ($\downarrow$)\end{tabular} &
  \multicolumn{1}{c|}{\begin{tabular}[c]{@{}c@{}}PCD\\ ($\downarrow$)\end{tabular}} &
  \multicolumn{1}{c|}{\begin{tabular}[c]{@{}c@{}}Acc-NoP\\ ($\uparrow$)\end{tabular}} &
  \begin{tabular}[c]{@{}c@{}}MAE-NoP\\ ($\downarrow$)\end{tabular} &
  \multicolumn{1}{c|}{\begin{tabular}[c]{@{}c@{}}PCD\\ ($\downarrow$)\end{tabular}} &
  \multicolumn{1}{c|}{\begin{tabular}[c]{@{}c@{}}Acc-NoP\\ ($\uparrow$)\end{tabular}} &
  \begin{tabular}[c]{@{}c@{}}MAE-NoP\\ ($\downarrow$)\end{tabular} &
  \multicolumn{1}{c|}{\begin{tabular}[c]{@{}c@{}}PCD\\ ($\downarrow$)\end{tabular}} &
  \multicolumn{1}{c|}{\begin{tabular}[c]{@{}c@{}}Acc-NoP\\ ($\uparrow$)\end{tabular}} &
  \begin{tabular}[c]{@{}c@{}}MAE-NoP\\ ($\downarrow$)\end{tabular} \\ \hline
\multicolumn{1}{|l|}{Path-wise} &
  48.03 \scriptsize $\pm 5.34$ &
  \multicolumn{1}{c|}{64.10 \scriptsize $\pm 7.79$} &
  \multicolumn{1}{c|}{73.40 \scriptsize $\pm 8.09$} &
  0.40 \scriptsize $\pm 0.09$ &
  \multicolumn{1}{c|}{40.45 \scriptsize $\pm 5.34$} &
  \multicolumn{1}{c|}{70.70 \scriptsize $\pm 10.30$} &
  0.35 \scriptsize $\pm 0.15$ &
  \multicolumn{1}{c|}{556.39 \scriptsize $\pm 10.60$} &
  \multicolumn{1}{c|}{\textbf{20.00} \scriptsize $\mathbf{\pm 3.04}$} &
  2.70 \scriptsize $\pm 0.23$ \\ \hline
\multicolumn{1}{|l|}{Autoregressive} &
  33.20 \scriptsize $\pm 6.16$ &
  \multicolumn{1}{c|}{45.52 \scriptsize $\pm 12.75$} &
  \multicolumn{1}{c|}{70.40 \scriptsize $\pm 11.05$} &
  0.38 \scriptsize $\pm 0.14$ &
  \multicolumn{1}{c|}{46.70 \scriptsize $\pm 8.06$} &
  \multicolumn{1}{c|}{81.40 \scriptsize $\pm 8.13$} &
  0.22 \scriptsize $\pm 0.09$ &
  \multicolumn{1}{c|}{708.71 \scriptsize $\pm 49.98$} &
  \multicolumn{1}{c|}{\textbf{15.56} \scriptsize $\mathbf{\pm 8.24}$} &
  \textbf{2.26} \scriptsize $\mathbf{\pm 0.39}$ \\ \hline
\multicolumn{1}{|l|}{Point-wise} &
  \textbf{6.76} \scriptsize $\mathbf{\pm 0.29}$ &
  \multicolumn{1}{c|}{7.45 \scriptsize $\pm 0.42$} &
  \multicolumn{1}{c|}{94.00 \scriptsize $\pm 3.11$} &
  0.09 \scriptsize $\pm 0.03$ &
  \multicolumn{1}{c|}{9.40 \scriptsize $\pm 0.47$} &
  \multicolumn{1}{c|}{93.15 \scriptsize $\pm 4.16$} &
  0.07 \scriptsize $\pm 0.04$ &
  \multicolumn{1}{c|}{\textbf{216.07} \scriptsize $\mathbf{\pm 17.85}$} &
  \multicolumn{1}{c|}{\textbf{17.22} \scriptsize $\mathbf{\pm 7.61}$} &
  \textbf{2.28} \scriptsize $\mathbf{\pm 0.24}$ \\ \hline\hline 
\multicolumn{1}{|l|}{\ours w/out AP2S} &
  7.79 \scriptsize $\pm 0.34$ &
  \multicolumn{1}{c|}{7.18 \scriptsize $\pm 0.30$} &
  \multicolumn{1}{c|}{\textbf{98.11} \scriptsize $\mathbf{\pm 0.74}$} &
  \textbf{0.05} \scriptsize $\mathbf{\pm 0.01}$ &
  \multicolumn{1}{c|}{11.27 \scriptsize $\pm 1.02$} &
  \multicolumn{1}{c|}{\textbf{97.00} \scriptsize $\mathbf{\pm 1.31}$} &
  \textbf{0.03} \scriptsize $\mathbf{\pm 0.02}$ &
  \multicolumn{1}{c|}{\textbf{220.66} \scriptsize $\mathbf{\pm 18.22}$} &
  \multicolumn{1}{c|}{\textbf{20.00} \scriptsize $\mathbf{\pm 4.69}$} &
  \textbf{2.30} \scriptsize $\mathbf{\pm 0.26}$ \\ \hline
\multicolumn{1}{|l|}{\ours} &
  \textbf{6.52} \scriptsize $\mathbf{\pm 0.28}$ &
  \multicolumn{1}{c|}{\textbf{6.83} \scriptsize $\mathbf{\pm 0.26}$} &
  \multicolumn{1}{c|}{\textbf{97.50} \scriptsize $\mathbf{\pm 0.85}$} &
  \textbf{0.05} \scriptsize $\mathbf{\pm 0.01}$ &
  \multicolumn{1}{c|}{\textbf{7.43} \scriptsize $\mathbf{\pm 0.35}$} &
  \multicolumn{1}{c|}{\textbf{98.10} \scriptsize $\mathbf{\pm 1.10}$} &
  \textbf{0.02} \scriptsize $\mathbf{\pm 0.01}$ &
  \multicolumn{1}{c|}{248.19 \scriptsize $\pm 41.39$} &
  \multicolumn{1}{c|}{\textbf{17.78} \scriptsize $\mathbf{\pm 12.51}$} &
  \textbf{2.32} \scriptsize $\mathbf{\pm 0.41}$ \\ \hline
\end{tabular}%
}
\end{table*}

\subsection{Results: Network Predictions}
\label{sec:results_network_predictons}
We evaluate the performance of models trained individually on each category, while keeping the same hyperparameters.
This allows us to study the robustness of our design choices to variations in data distributions.
The categories are considered in order of growing complexity of object shapes and path patterns: Cuboids, Windows, Shelves and Containers. 

\medskip
\noindent \textbf{Main qualitative results.}
We report the qualitative results of raw network predictions on a subset of test instances in Fig.~\ref{fig:qualitatives_segments_prediction}.
The results from the \emph{Path-wise} baseline indicate that one-shot methods inspired by object detection~\cite{redmon2016yolo,carion2020detr,cheng2021maskformer} provide a strong basis for addressing unstructured output paths.
Although promising, the major limitation of this approach stems from the explicit computation of the loss between high-dimensional paths: the curse of dimensionality yields inaccurate predictions over long-time horizons. This can be noted by observing the degenerate raster patterns for Cuboids and Containers. 

The \emph{Autoregressive} approach aims at counteracting the described  phenomenon, by reducing the dimensionality of the network output at the cost of additional forward passes.
However, this baseline exhibits significant compounding errors, often predicting the wrong number of straight passes when attempting to match the long raster paths of the Cuboids category.
These findings highlight the importance of coping with the long-horizon nature of output paths alongside the other challenges of OCMG tasks.

Both \emph{Point-wise} and \ours approach the task by focusing on local path patterns and generating sets of poses or path segments, respectively. 
This novel paradigm shows significantly stronger generalization capabilities across all object categories.
Still, the Point-wise baseline provides predictions that fail to capture detailed path structures---\eg, see the top face of the Cuboid, and the inner paths of the Shelf.
\ours successfully tackles this problem by introducing (1) segment-wise Chamfer Distance terms, and (2) the Asymmetric Point-to-Segment Curriculum (AP2S). Notice that both enable \ours's state-of-the-art performance: the paths obtained by \ours without AP2S exhibit errors similar to those of the Point-wise baseline. In other words, incorporating both point-wise and segment-wise terms in $\mathcal{L}_{p2s}$ yields better performance than minimizing either one alone (see Appendix~\ref{appendix:pointtosegment_ablation} for a thorough ablation of our loss function).

Finally, in Fig.~\ref{fig:intermediate_results} we present an illustration of \ours's network predictions as the training progresses. The picture highlights the effect of the asymmetric curriculum: the model initially learns to generate sparse individual poses close to the target object surface; then, local structure is gradually promoted and smooth segment predictions are obtained.

\medskip
\noindent \textbf{Main quantitative results.}
The quantitative results of our experimental analysis are presented in Tab.~\ref{tab:quantitative_all_metrics}, considering all the metrics defined in Sec.~\ref{sec:eval_metrics}.
It can be observed that the novel mask predictions paradigm is the most effective across all object categories.
The Path-wise and Autoregressive baselines attempt to learn a confidence score for each path, but underperform in terms of NoP metrics.
Instead, Point-wise and \ours predict non-mutually exclusive masks over their own predictions to distinguish among different paths, a strategy that leads to significantly better NoP metrics at inference time.
Notably, these findings align with those on the predicted number of instances from the Panoptic Segmentation literature, where MaskFormer~\cite{cheng2021maskformer} improves over DETR~\cite{carion2020detr}.

Finally, we analyze the results on the Containers category in greater detail. Here, the models are trained on only 70 data samples of heterogeneous object shapes, and tested on 18 unseen instances.
This is a challenging setting often encountered in real-world industrial scenarios where the training set size is limited, while strong generalization capabilities are required.
All methods exhibit low performance across the evaluated metrics, with high variance across independent training repetitions.
Although Point-wise achieves PCD scores comparable to our method, we observe that the former leads to globally sparse predictions that lack local consistency (cf. Fig.~\ref{fig:qualitatives_segments_prediction}).
We attribute this behavior to the properties of the Chamfer Distance, which is well-known for its insensitivity to mismatched local density~\cite{densityawarecd}.
Yet, \ours outperforms the other baselines in terms of PCD also on the Containers category.
Generalization in such data-scarce conditions can be further improved by leveraging pre-trained models or incorporating additional data across different categories, as discussed in Sec.~\ref{sec:results_generalization}.

\begin{figure}[] 
    \centering
    \includegraphics[width=\linewidth]{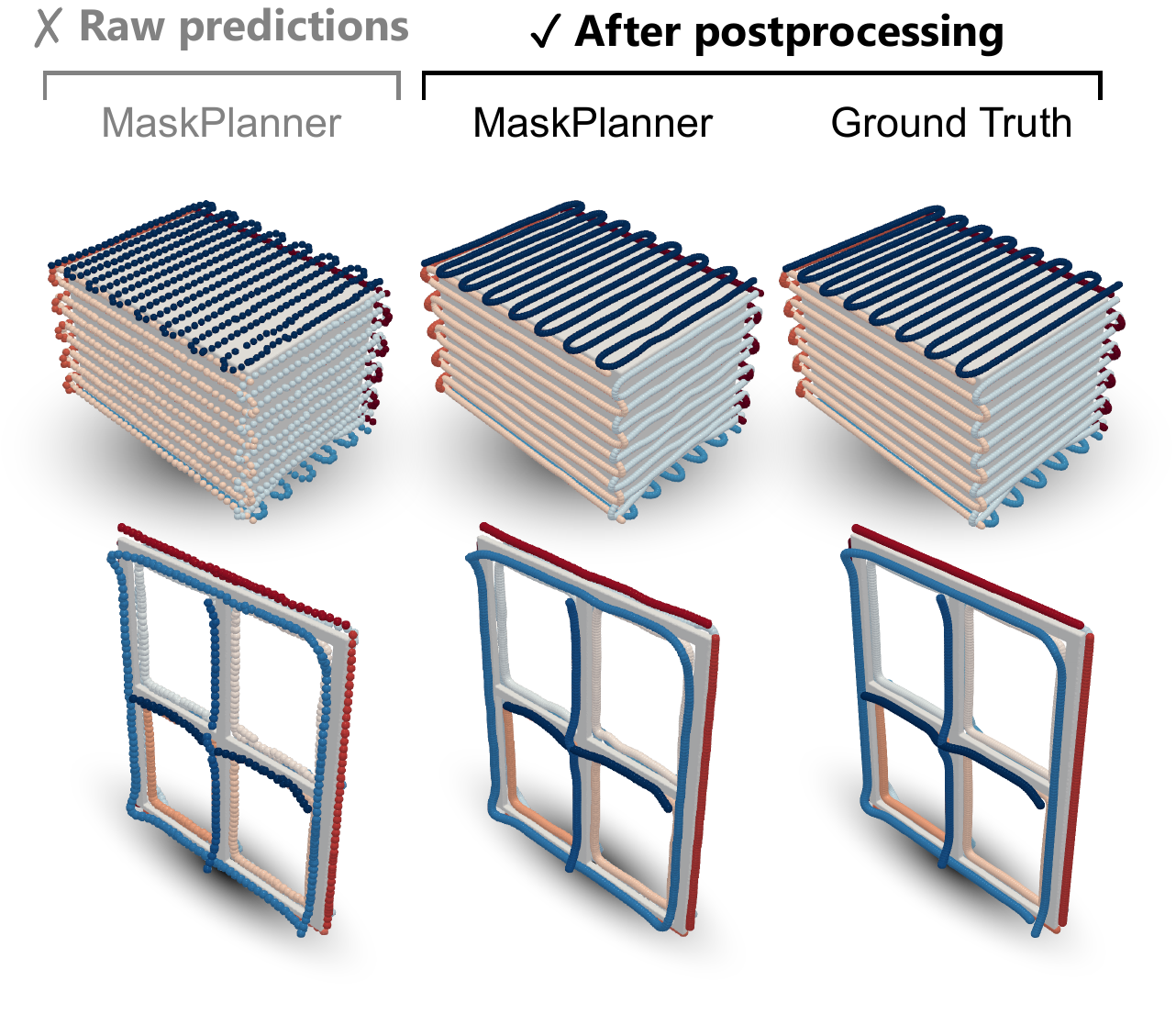}
    \vspace{-28pt}
    \caption{Final paths when postprocessing is applied to segment predictions on Cuboids and Windows.}
    \vspace{-6pt}
    \label{fig:example_postprocess}
\end{figure}

\noindent \textbf{Postprocessing.}
The postprocessing step described in Sec.~\ref{sec:postprocessing} aims at preparing the generated paths for execution on robotic systems.
Its practical effect on concatenating and smoothing the predicted segments can be observed in Fig.~\ref{fig:example_postprocess}.
We remark that the choice of the segments' length $\lambda$ can affect the postprocessing, potentially leading to unfeasible paths.

\begin{figure}[]
    \centering
    \includegraphics[width=\linewidth]{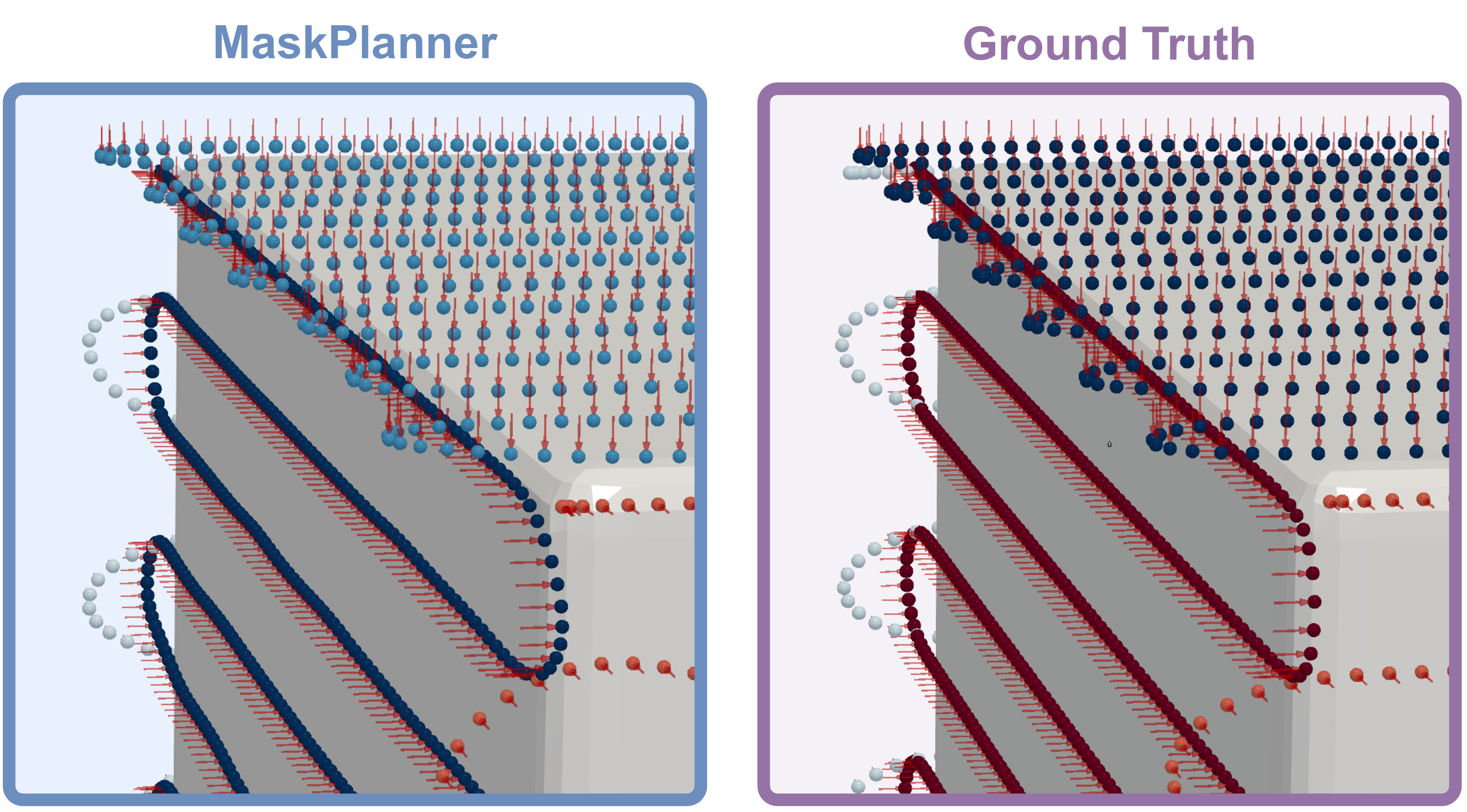}
    \caption{
    Close up \ours results after postprocessing: both pose locations and orientations (red arrows) are effectively learned.
    }
    \label{fig:closeup}
    \vspace{-6pt}
\end{figure}

\subsection{Results: Generalization capabilities}
\label{sec:results_generalization}
By formalizing OCMG as a supervised learning task we are able to address key challenges commonly encountered in real-world industrial scenarios, such as the need to adapt to new data as it becomes available.
To this end, data-driven approaches offer several advantages over ad-hoc heuristics or traditional optimization-based methods. 
For example, models can be efficiently re-trained without requiring process re-engineering, and it is possible to leverage pre-trained models to improve performance in cases of limited data or time constraints. In the following, we analyze how different training configurations influence the generalization capabilities of our method.

\begin{figure}[tb]
\centering
\begin{subfigure}[b]{\linewidth}
   \centering
   \includegraphics[width=0.55\linewidth]{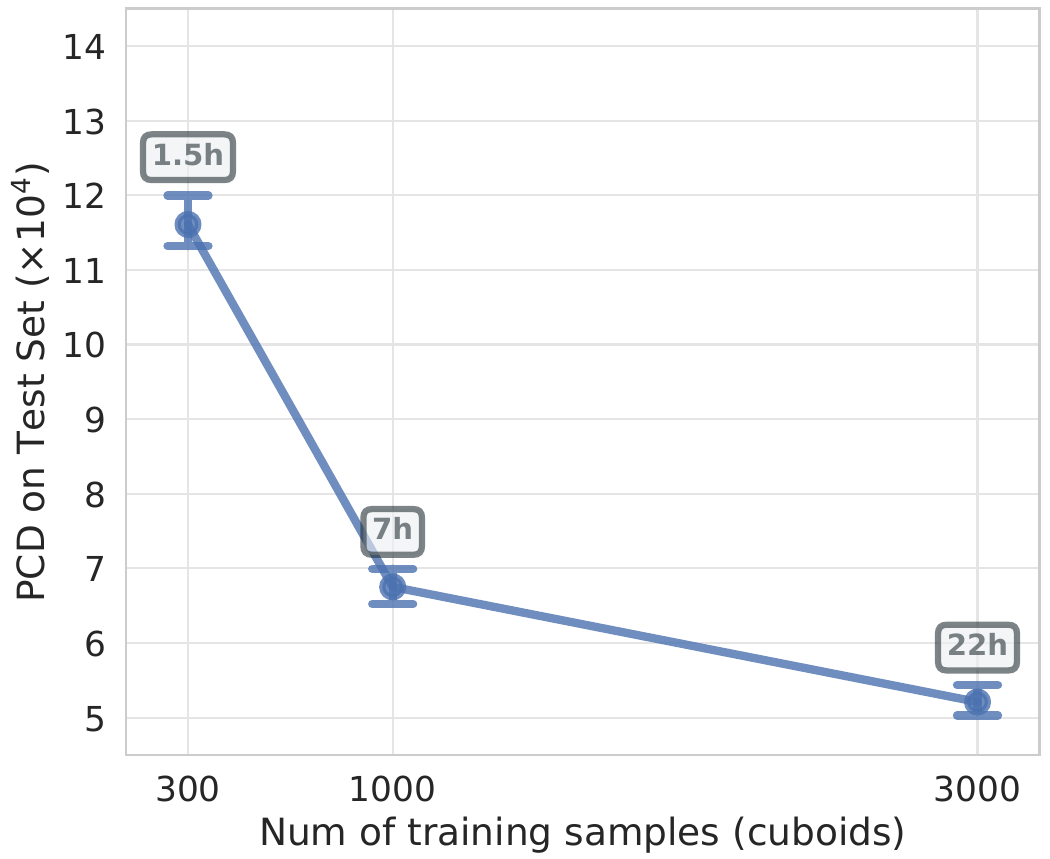}
   \vspace{-4pt}
   \caption{}
   \label{fig:Ng1} 
\end{subfigure}
\begin{subfigure}[b]{\linewidth}
    \centering
   \includegraphics[width=\linewidth]{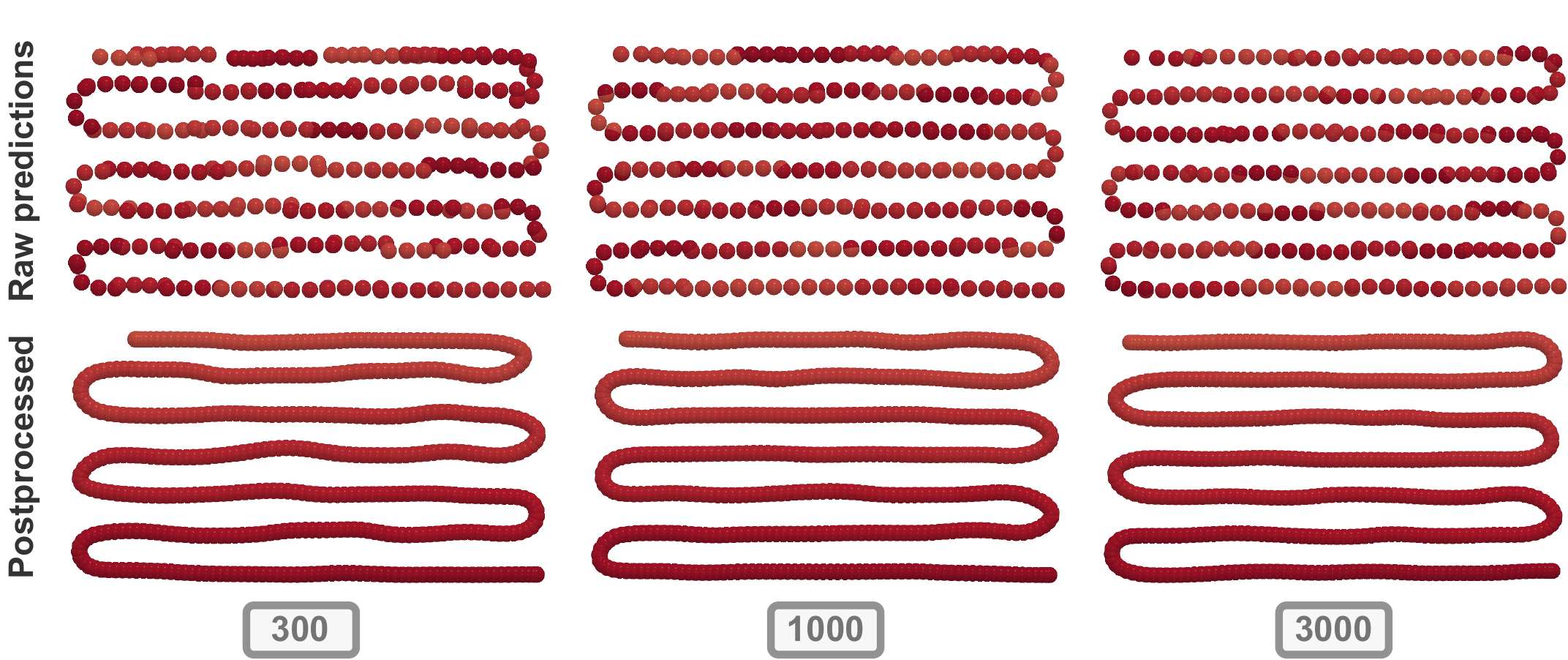}
   \caption{}
   \label{fig:Ng2}
\end{subfigure}

\caption{(a) Generalization performance for varying amounts of training data, on the Cuboids category (5 seeds, average training time is displayed as text). For each seed, a new model is trained on different number of samples and tested on the same, fixed test set. (b) A representative predicted path at test time is depicted for each of the three configurations, before (top) and after (bottom) postprocessing.}
\label{fig:n_samples_scaling}
\end{figure}

\noindent \textbf{Sensitivity to training set size.}
We investigate the performance trend of \ours when scaling up the amount of available training data. Specifically, we focus on the Cuboids category and report our findings for a model trained on 300, 1000, or 3000 samples.
The results in Fig.~\ref{fig:n_samples_scaling} show that training on 300 Cuboids yields high PCD (\ie, high error) and paths with visible irregularities. These issues are largely mitigated when considering 1000 training samples and further improve when increasing the training set size to 3000. 

\begin{figure}[] 
    \centering
    \includegraphics[width=\linewidth]{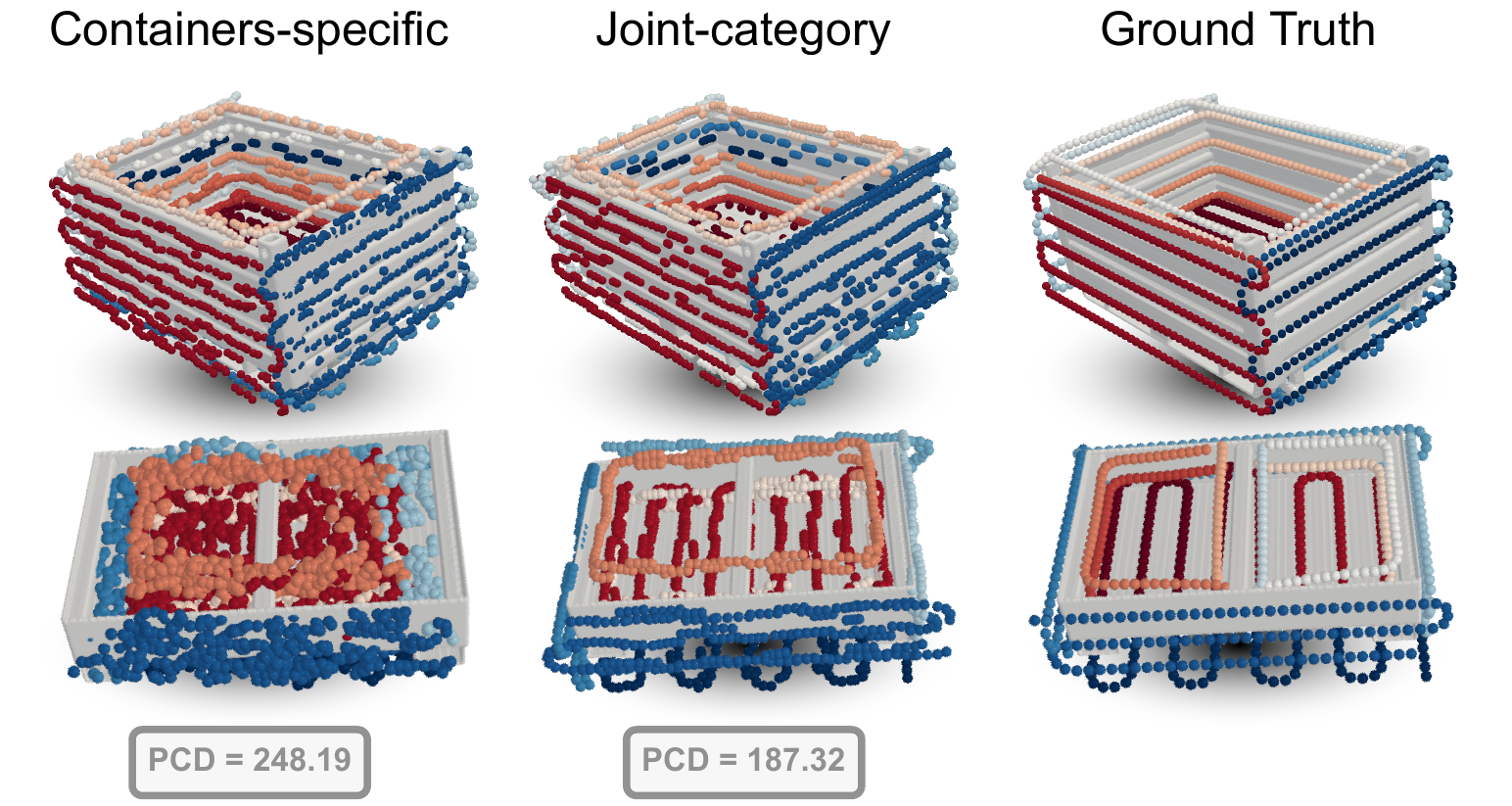}
    \caption{Performance comparison on two test samples between a Containers-specific model vs.~a model trained on all categories combined (joint-category). Quantitative metrics refer to average over all 18 container test samples.}
    \label{fig:jointtraining}
\end{figure}


\noindent \textbf{Joint-category training.}
We extend the aforementioned analysis by studying the effect of jointly training \ours on multiple object categories with the goal of answering the following question:~\emph{``Does augmenting the training set with data from a different distribution but same task yield higher predictive performance?''}
We illustrate the results in Fig.~\ref{fig:jointtraining}, where the basic model trained only on 70 Containers is compared to a model jointly trained  on Cuboids, Windows, Shelves, and Containers.
Notably, the latter achieves significantly lower PCD on unseen Containers. It is qualitatively evident that the Containers-specific model is unable to preserve segments structure for novel shapes, showing more scattered predictions over the object surface. 
We conclude that experiencing higher shape and path diversity during training improves the model's generalization ability.

\begin{figure}[]
    \centering
    \includegraphics[width=\linewidth]{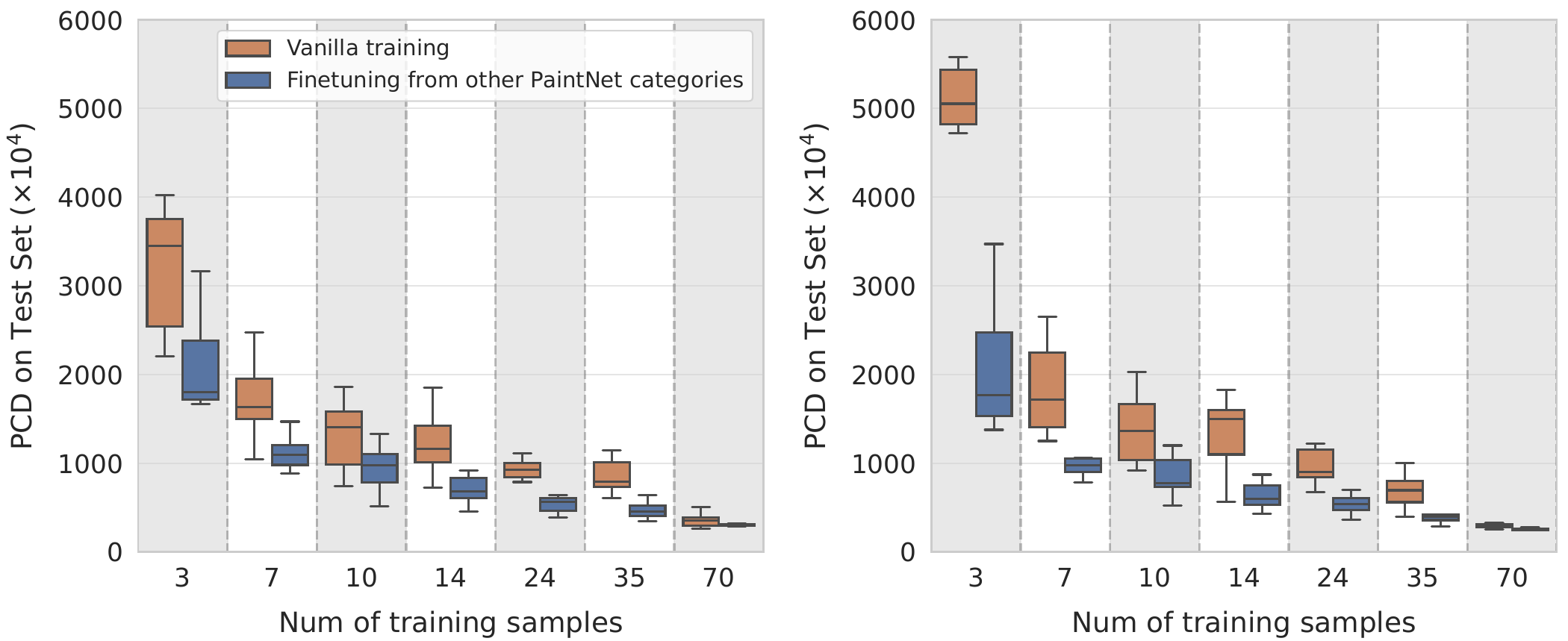}
    \caption{
    Few-shot: a model jointly pre-trained on Cuboids, Shelves, and Windows generalizes better when finetuned on a subset of Containers. Results on test set after (left) 600 and (right) 1200 training epochs; 10 repetitions.
    }
    \label{fig:few_shot}
\end{figure}

\begin{figure}[]
    \centering
    \includegraphics[width=\linewidth]{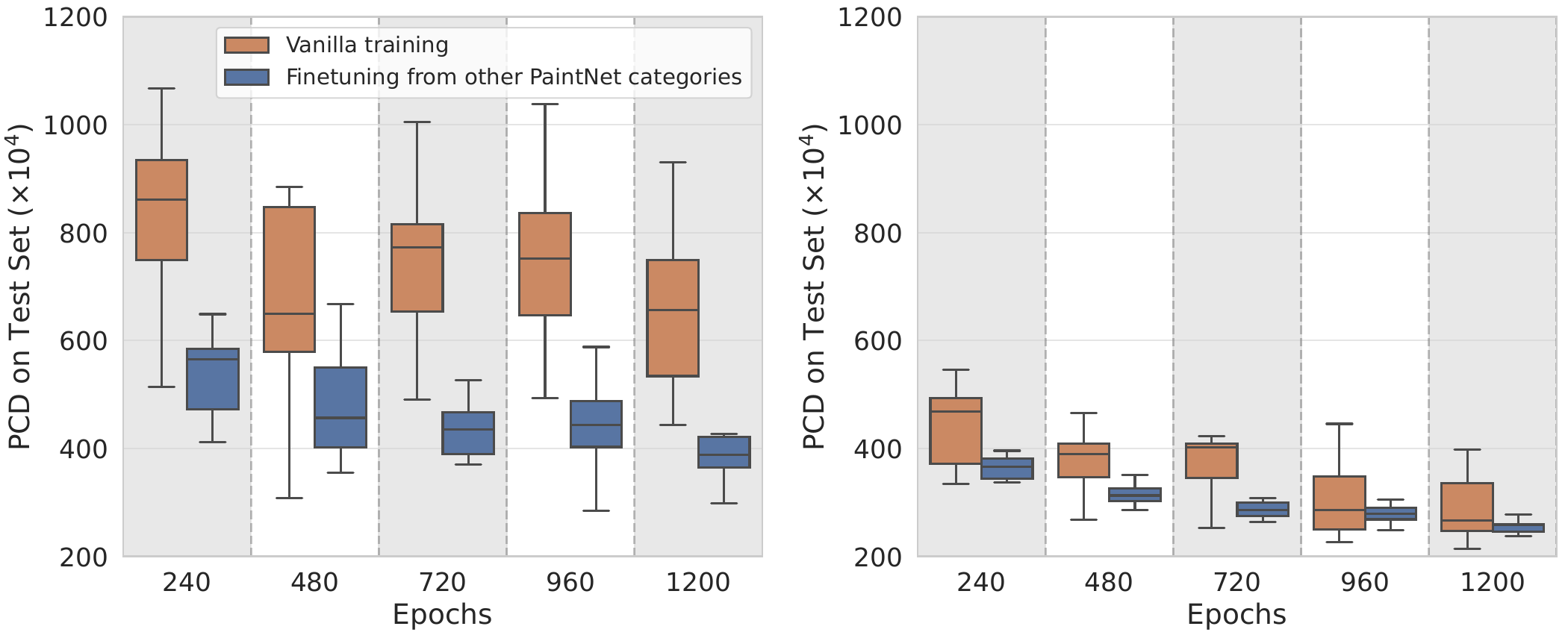}
    \caption{
    Convergence speed: a model jointly pre-trained on Cuboids, Shelves, and Windows leads to faster convergence when finetuned on Containers. Training with (left) 50\% and (right) 100\% of available containers; 10 repetitions.
    }
    \label{fig:conv_speed}
\end{figure}

\noindent \textbf{Finetuning on a novel category.}
Finally, we consider a knowledge transfer scenario in which \ours pre-trained on Cuboids, Shelves, and Windows is subsequently finetuned on the Containers category.
We compare the resulting performance with that obtained by \ours in the vanilla setting which leverages the task of shape classification on ModelNet as pre-training for backbone initialization. In particular, we inspect how performances vary across data and time constraints. The results in Fig.~\ref{fig:few_shot} show that exploiting pre-trained models on task-specific data provides an advantage over the vanilla strategy in the few-shot scenario with scarce availability of training data.
Similar conclusions can be drawn from the results in Fig.~\ref{fig:conv_speed}, where we analyze the convergence speed of the two approaches in terms of training epochs.

These findings provide a clear indication of the potential of supervised learning for future OCMG applications.
Indeed, the real-time inference capabilities of these models combined with an increasing amount of data can drastically reduce robot programming times.

\subsection{Results: Spray Painting in Simulation}
\label{sec:results_spray_painting}
\ours and the proposed baselines aim at solving the spray painting task purely from expert data, \ie achieving high paint coverage despite not being explicitly optimized for a task-specific objective.
To evaluate their capabilities on the downstream task, we run a spray painting simulation by placing the gun nozzle at each 6D waypoint predicted by the network (as displayed in Fig.~\ref{fig:qualitatives_segments_prediction}), at discrete time steps.
By doing so, the resulting paint coverage is invariant under permutation of the predicted waypoints and their arrangement into separate paths.
Thus, we can fairly compare all the methods regardless of their point, segment, or path output structure. 
We execute the evaluation using a proprietary painting simulator, although similar tools would be equally suitable~\cite{Andulkar_Incremental_2015}.

The results in Tab.~\ref{tab:quantitative_coverage} show two main trends: \ours achieves (1) substantially higher paint coverage on average, and (2) exhibits the lowest variance across test samples on all object categories.
Notably, over $99\%$ of the object meshes are covered for test instances of Cuboids, Windows, and Shelves when executing \ours's predictions.

\begin{figure*}[!t]
    \centering
    \includegraphics[width=\linewidth]{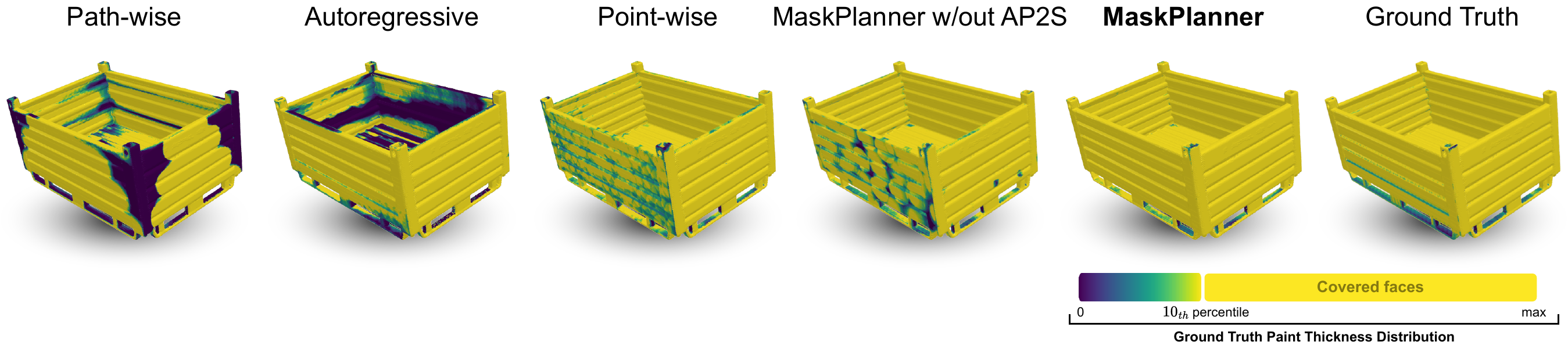}
    \vspace{-16pt}
    \caption{
    Qualitative paint coverage results in simulation on a representative test instance of the Containers category.
    }
    \label{fig:qualitatives_pc_containers}
\end{figure*}

Results on the Containers category provide further insights into the performance of \ours.
In line with the PCD results in Tab.~\ref{tab:quantitative_all_metrics}, \ours attains a significant improvement in terms of paint coverage compared to the Path-wise and Autoregressive baselines.
We report qualitative results for a representative Container object in Fig.~\ref{fig:qualitatives_pc_containers}.
Here, it is evident that Path-wise and Autoregressive predictions display issues in managing high dimensional paths or diverge due to compounding errors---as mentioned in Sec.~\ref{sec:results_network_predictons} and highlighted in Fig.~\ref{fig:qualitatives_segments_prediction}. 
As a result, large portions of the surface remain uncovered. 

\begin{table}[]
\centering
\caption{
Quantitative paint coverage results, with Mean and St.dev. across instances of the test set. Mean values that are \emph{not} statistically significantly worse than any other are marked in bold ($\alpha {=} 0.05$).
}
\label{tab:quantitative_coverage}
\resizebox{\columnwidth}{!}{%
\def\arraystretch{1.25}%
\begin{tabular}{lcccc}
 &
  \multicolumn{4}{c}{Paint Coverage \% ($\uparrow$)} \\ \cline{2-5} 
\multicolumn{1}{l|}{} &
  \multicolumn{1}{c|}{Cuboids} &
  \multicolumn{1}{c|}{Windows} &
  \multicolumn{1}{c|}{Shelves} &
  \multicolumn{1}{c|}{Containers} \\ \hline
\multicolumn{1}{|l|}{Path-wise} &
  \multicolumn{1}{c|}{77.54 \scriptsize $\pm 11.89$} &
  \multicolumn{1}{c|}{81.50 \scriptsize $\pm 7.83$} &
  \multicolumn{1}{c|}{91.70 \scriptsize $\pm 3.58$} &
  \multicolumn{1}{c|}{58.37 \scriptsize $\pm 19.31$} \\ \hline
\multicolumn{1}{|l|}{Autoregressive} &
  \multicolumn{1}{c|}{88.49 \scriptsize $\pm 5.73$} &
  \multicolumn{1}{c|}{88.20 \scriptsize $\pm 6.17$} &
  \multicolumn{1}{c|}{89.10 \scriptsize $\pm 2.97$} &
  \multicolumn{1}{c|}{53.29 \scriptsize $\pm 17.06$} \\ \hline
\multicolumn{1}{|l|}{Point-wise} &
  \multicolumn{1}{c|}{\textbf{94.02} \scriptsize $\mathbf{\pm 8.16}$} &
  \multicolumn{1}{c|}{90.34 \scriptsize $\pm 7.30$} &
  \multicolumn{1}{c|}{92.34 \scriptsize $\pm 7.79$} &
  \multicolumn{1}{c|}{\textbf{88.29} \scriptsize $\mathbf{\pm 13.67}$} \\ \hline\hline
\multicolumn{1}{|l|}{\ours w/out AP2S} &
  \multicolumn{1}{c|}{\textbf{99.87} \scriptsize $\mathbf{\pm 0.40}$} &
  \multicolumn{1}{c|}{\textbf{99.37} \scriptsize $\mathbf{\pm 0.73}$} &
  \multicolumn{1}{c|}{98.86 \scriptsize $\pm 2.13$} &
  \multicolumn{1}{c|}{\textbf{88.63} \scriptsize $\mathbf{\pm 17.16}$} \\ \hline
\multicolumn{1}{|l|}{\ours} &
  \multicolumn{1}{c|}{\textbf{99.82} \scriptsize $\mathbf{\pm 0.48}$} &
  \multicolumn{1}{c|}{\textbf{99.36} \scriptsize $\mathbf{\pm 0.62}$} &
  \multicolumn{1}{c|}{\textbf{99.76} \scriptsize $\mathbf{\pm 0.60}$} &
  \multicolumn{1}{c|}{\textbf{90.94} \scriptsize $\mathbf{\pm 12.93}$} \\ \hline
\end{tabular}%
}
\vspace{18pt}
\end{table}

\subsection{Results: Spray Painting Validation in the Real World}
\label{sec:real_world_exps}
We conclude our experimental evaluation by demonstrating the successful application of \ours to a real-world industrial robotic system on the production line. 
Our validation aims to answer the following questions:
\begin{itemize}
    \item \emph{Are paths predicted by \ours kinematically and dynamically feasible for direct execution on a real robot?}
    \item \emph{Does near-complete coverage in simulation translate to near-complete coverage of real objects?}
\end{itemize}
To this end, two representative object instances are drawn from the test set of Cuboids and Windows and hand-crafted by experts with their nominal dimensions.
Note that, by doing so, the object point-cloud is readily available from our dataset and no domain shift between synthetic and real data is experienced by the network at inference time.

\begin{figure*}[!t]
    \centering
    \includegraphics[width=\linewidth]{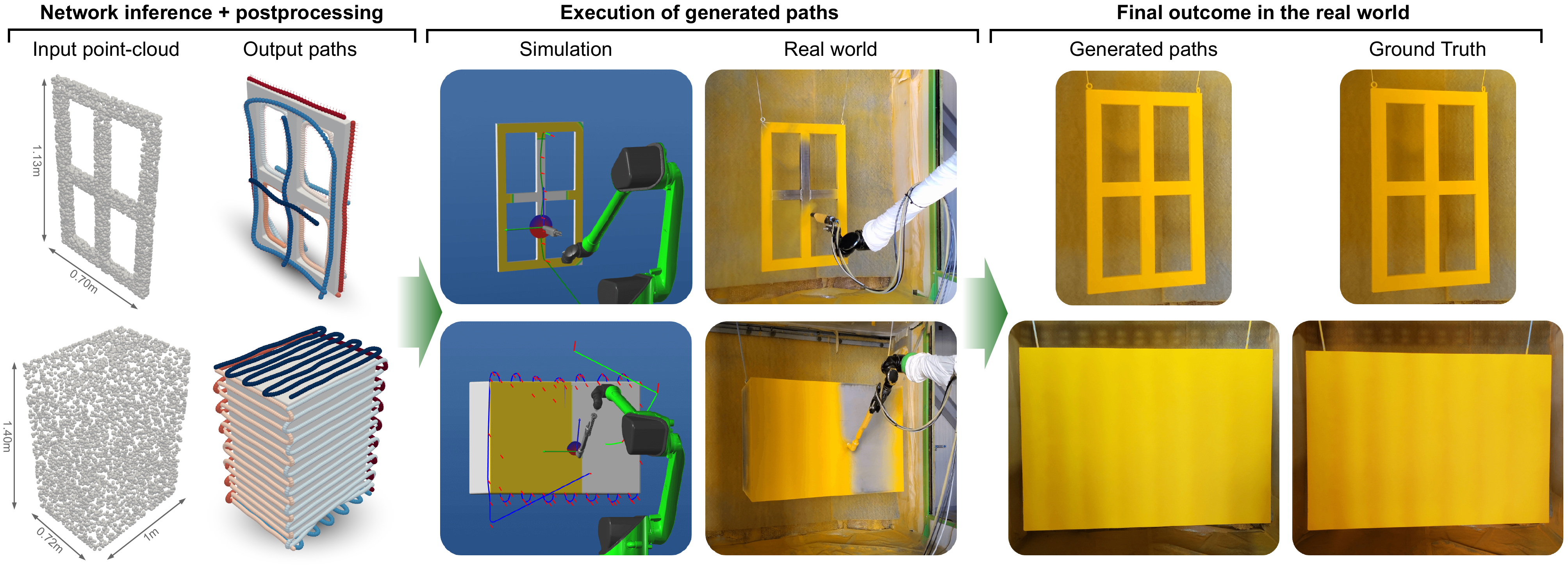}
    \vspace{-12pt}
    \caption{
    Real-world validation of \ours on two test objects. A set of long-horizon paths is inferred given the object point clouds through a single forward pass ($100ms$) and a postprocessing step ($100ms$). Then, paths are checked for kinematic and dynamic feasibility in simulation and later executed on the real setup.
    The final paint result on the real objects is effectively equivalent to that produced by ground truth paths.
    }
    \label{fig:realworld_validation_results}
\end{figure*}

We design the experiment for powder coating, a widely adopted spray painting technique where paint is applied electrostatically to the object and subsequently cured under heat.
We refer the reader to Fig.~\ref{fig:realworld_validation_results} for an illustration of the full real-world validation pipeline.

\noindent \textbf{Inference and execution.}
We deploy the same category-specific models whose results have been discussed in Sec.~\ref{sec:results_network_predictons}  
to predict output segments for the Cuboid and Window instances displayed in Fig.~\ref{fig:realworld_validation_results} (left).
Predicted segments are then automatically concatenated through our postprocessing step to generate output paths.
The entire path generation pipeline only takes $200ms$ for each input object, requires a single forward pass of the network, and results in six paths for the Cuboid (total length of 83 meters) and 14 paths for the Window (total length of 17 meters).
The generated paths can be then executed on real robots by employing any approach of choice, such as tracking waypoints in Cartesian space.
In our setting, we apply the Ramer–Douglas–Peucker algorithm---with thresholds of $1cm$ for translational coordinates and $15deg$ for rotational coordinates---and interpolate the subsampled waypoints through smooth junctions that preserve acceleration limits while targeting a desired velocity.
In particular, we set the target velocity to $25cm/s$.
A dedicated software automatically detects and manages reachability issues: in our physical setup it occasionally applied minor adjustments to the orientation normals of the paths (\eg, for the four inner and four outer edges of the window).

Furthermore, motivated by the symmetries of the considered objects, we avoid painting the back side of the window and cuboid without loss of generality.
We execute paths in a random permutation, and reset the robot to a predefined starting configuration between each path execution.
Both the generated and the ground-truth paths are then executed on a real
Efort GR-680 6-DoF specialized painting robot, with their final outcome depicted on the right side of Fig.~\ref{fig:realworld_validation_results}.

\noindent \textbf{Results and discussion.}
All paths generated by \ours were executed successfully on real hardware.
Paths were found to be dynamically feasible for direct execution, and no collisions occurred at any time.
Importantly, the final spray painting outcome achieved with \ours trajectories proved indistinguishable from that of ground-truth expert trajectories across both objects (cf. Fig.~\ref{fig:realworld_validation_results}).
Remind that \ours includes no explicit paint optimization subroutine, rendering the final result a ground-breaking demonstration of its potential for addressing OCMG tasks through a pure data-driven perspective.
Notably, we observe slight uneven paint deposition across both ground truth and predicted paths on the real Cuboid (see the vertical stripe trend in Fig.~\ref{fig:realworld_validation_results}).
This effect can be mitigated by domain experts with further application-specific trajectory optimization techniques~\cite{gleeson2022generating}, or manual tuning of the target velocity for different paint types.
Nevertheless, our real-world validation experiment demonstrates the effectiveness of \ours in quickly generating expert-level path patterns for previously unseen object instances, while requiring no domain knowledge and remaining applicable to a variety of object-centric motion generation tasks.

\section{Conclusions}
\label{sec:conclusions}
In this paper, we address the core problem of robot motion generation conditioned on free-form 3D objects. Particularly, we formalize the Object-Centric Motion Generation (OCMG) problem setting, aiming to unify several robotic applications under a common framework for the prediction of long-horizon, unstructured paths.
To tackle this problem, we introduce \ours, a novel deep learning method capable of inferring smooth and accurate paths directly from expert data.
In simple terms, \ours breaks down motion generation into the joint prediction of local path segments and probability masks, followed by a postprocessing step that concatenates the segments predicted within the same mask.
Our approach demonstrates generalization across both convex and concave 3D objects (Cuboids, Windows, Shelves), after being trained for only 6 hours on 800 samples on a single NVIDIA GeForce RTX 4080 GPU.
We validate \ours in the context of robotic spray painting, demonstrating its ability to achieve near-complete paint coverage on previously unseen objects, both in simulation and in real-world experiments.
Our method remains task-agnostic and makes minimal assumptions on both the object geometry and the output path patterns, paving the way for future applications of deep learning to OCMG tasks beyond spray painting---such as welding, sanding, cleaning, or visual inspection.
In fact, our findings showcase the potential of data-driven methods to achieve solutions to OCMG tasks that are scalable, generalizable, and cheap to deploy: models can (1) be pre-trained with data from similar tasks, (2) boosted in performance when more samples become available, (3) leverage heuristics-based demonstrations, and (4) be deployed at 10Hz on large objects.
Remarkably, we believe this approach can shine even for high-precision tasks when used in combination with task-specific trajectory optimization techniques that leverage \ours to promptly generate good starting solutions for novel objects.
Furthermore, \ours enables the generation of unstructured paths from data modalities beyond 3D objects by replacing the feature encoder, opening new directions of work for applications such as agricultural coverage path planning from aerial images and multi-UAV search-and-rescue missions.

\noindent\textbf{Limitations.}  Our current pipeline relies on a simple postprocessing strategy to concatenate predicted segments, which is not able to recover from erroneous predictions and may lead to inaccurate paths in such cases. Learning-based modules for concatenation will be investigated in future work to be robust to misplaced segment predictions.
We also reckon that the choice of the segment length $\lambda{=}4$ may not work equally well across different tasks and objects.
This hyperparameter should be tuned accordingly to learn accurate isolated patterns that generalize well across the object surface.
Alternatively, tuning the granularity of the sampled waypoints from the expert trajectories to achieve the desired degree of sparsity in the predicted poses would serve a similar purpose.
Finally, our practical implementation simplifies gun orientations to unit vectors (from 3-DoF to 2-DoF), making the sole assumption that the task is invariant to gun rotations around the approach axis.
Future work can extend the pipeline to the prediction of full orientation representations---\eg, Euler angles---to tackle tasks where the additional degree of freedom is necessary, such as motion generation for object grasping.

\section*{Acknowledgments}
The authors acknowledge the EFORT group's support, which provided domain knowledge, object meshes, expert trajectory data, and access to specialized painting robot hardware for our real-world experimental evaluation.
This study was carried out within the FAIR - Future Artificial Intelligence Research and received funding from the European Union Next-GenerationEU (PIANO NAZIONALE DI RIPRESA E RESILIENZA (PNRR) – MISSIONE 4 COMPONENTE 2, INVESTIMENTO 1.3 – D.D. 1555 11/10/2022, PE00000013). This manuscript reflects only the authors’ views and opinions, neither the European Union nor the European Commission can be considered responsible for them.

{\appendices
\section{Asymmetric Point-to-Segment Curriculum}
\label{appendix:pointtosegment_ablation}

When training \ours to minimize the proposed loss $\mL_{p2s}$ (see Sec.~\ref{sec:segments_prediction}), various weighting schemes can be applied to the point-wise and segment-wise Chamfer Distance (CD) terms.
In principle, an ideal learning algorithm would drive both point-wise and segment-wise matching simultaneously.
In practice, however, gradient-based optimization can converge to substantially different local optima---\eg while the Chamfer Distance is computationally efficient, it's known to be sensitive to outliers and insensitive to local mismatches in density~\cite{densityawarecd}.
In our ablation study (see Tab.~\ref{tab:loss_ablation_v2}), we evaluate four main weighting configurations:
\noindent\begin{itemize}
    \item \textbf{\ours w/out AP2S}: a baseline with a fully segment-wise loss function that leverages no auxiliary point-wise CD terms. This weighting scheme also resembles the loss function introduced in~\cite{tiboni2023paintnet}.
    
    \item \textbf{(1)~Asymmetric}: keeps a segment-wise forward CD term, but uses a point-wise loss function for the backward CD term.

    \item \textbf{(2)~P2S curriculum}: uses a coarse-to-fine schedule to progressively assign more weight to segment-wise predictions, but does so symmetrically for both forward and backward terms.

    \item \textbf{(1)~+~(2)~\ours}: our full asymmetric point-to-segment curriculum that starts training with a higher weight on the backward point-wise term and gradually balances both point-wise and segment-wise backward terms. The forward term is fully segment-wise throughout training.
\end{itemize}

Quantitative results show that \ours converges to lower PCD scores when both asymmetric CD terms and a gradual point-to-segment curriculum are included.
In particular, we observe that including the computation of auxiliary point-wise terms only helps if used in combination with the AP2S curriculum.
A variety of additional configurations have also been tried, but led to no improvements.
Moreover, we remind that na\"ively optimizing for fully point-wise CD terms yields qualitative sparse predictions that fail in capturing detailed path structures (see Point-wise baseline in Fig.~\ref{fig:qualitatives_segments_prediction}).
In turn, we conclude that the AP2S curriculum is crucial to promote effective convergence while ensuring smoothness and local consistency in the final predictions.

\begin{table}[]
\centering
\caption{PCD on the test set of all object categories when models are trained on $\mL_{p2s}$ with different weighting configurations (10 seeds). The ``$a\rightarrow b$" notation indicates the value varies from $a$ to $b$ across training.}
\label{tab:loss_ablation_v2}
\resizebox{\columnwidth}{!}{%
\def\arraystretch{1.25}%
\begin{tabular}{l|cl|cl|cl|cl|}
\cline{2-9}
 &
  \multicolumn{2}{c|}{\begin{tabular}[c]{@{}c@{}}\ours\\ w/out AP2S\end{tabular}} &
  \multicolumn{2}{c|}{\begin{tabular}[c]{@{}c@{}}(1)\\ Asymmetric\end{tabular}} &
  \multicolumn{2}{c|}{\begin{tabular}[c]{@{}c@{}}(2)\\ P2S Curr.\end{tabular}} &
  \multicolumn{2}{c|}{\textbf{\begin{tabular}[c]{@{}c@{}}(1) + (2)\\ \ours\end{tabular}}} \\ \cline{2-9} 
 &
  \multicolumn{2}{c|}{\begin{tabular}[c]{@{}c@{}}$w^f_p=0$\\ $w^f_s=1$\\ $w^b_p=0$\\ $w^b_s=1$\end{tabular}} &
  \multicolumn{2}{c|}{\begin{tabular}[c]{@{}c@{}}$w^f_p=0$\\ $w^f_s=1$\\ $w^b_p=1$\\ $w^b_s=0$\end{tabular}} &
  \multicolumn{2}{c|}{\begin{tabular}[c]{@{}c@{}}$w^f_p=100\rightarrow 1$\\ $w^f_s=0.01\rightarrow 1$\\ $w^b_p=100\rightarrow 1$\\ $w^b_s=0.01\rightarrow 1$\end{tabular}} &
  \multicolumn{2}{c|}{\begin{tabular}[c]{@{}c@{}}$w^f_p=0$\\ $w^f_s=1$\\ $w^b_p=100\rightarrow 1$\\ $w^b_s=0.01\rightarrow 1$\end{tabular}} \\ \hline
\multicolumn{1}{|l|}{Cuboids}    & \multicolumn{2}{c|}{7.79}            & \multicolumn{2}{c|}{11.81}  & \multicolumn{2}{c|}{9.24}   & \multicolumn{2}{c|}{\textbf{6.52}} \\ \hline
\multicolumn{1}{|l|}{Windows}    & \multicolumn{2}{c|}{7.18}            & \multicolumn{2}{c|}{12.55}  & \multicolumn{2}{c|}{11.89}  & \multicolumn{2}{c|}{\textbf{6.83}} \\ \hline
\multicolumn{1}{|l|}{Shelves}    & \multicolumn{2}{c|}{11.27}           & \multicolumn{2}{c|}{30.65}  & \multicolumn{2}{c|}{34.84}  & \multicolumn{2}{c|}{\textbf{7.43}} \\ \hline
\multicolumn{1}{|l|}{Containers} & \multicolumn{2}{c|}{\textbf{220.66}} & \multicolumn{2}{c|}{242.07} & \multicolumn{2}{c|}{240.19} & \multicolumn{2}{c|}{248.19}        \\ \hline
\end{tabular}%
}
\end{table}

} 

\bibliographystyle{IEEEtran}
\balance{}
\bibliography{bibliography}

\end{document}